\documentclass{article}

\usepackage{microtype}
\usepackage{graphicx}
\usepackage{subcaption}
\usepackage{booktabs} 
\usepackage[table]{xcolor} 
\usepackage{multirow} 
\usepackage{array}
\usepackage{hyperref}

\usepackage[preprint]{icml2026}

\usepackage{amsmath}
\usepackage{amssymb}
\usepackage{mathtools}
\usepackage{amsthm}

\usepackage[capitalize,noabbrev]{cleveref}

\theoremstyle{plain}

\theoremstyle{definition}

\theoremstyle{remark}

\usepackage[textsize=tiny]{todonotes}

\icmltitlerunning{Streaming-dLLM: Accelerating Diffusion LLMs via Suffix Pruning and Dynamic Decoding}
\definecolor{mytan}{RGB}{255, 250, 240}
\begin{document}

\twocolumn[
  \icmltitle{Streaming-dLLM: Accelerating Diffusion LLMs via\texorpdfstring{\\}{ }Suffix Pruning and Dynamic Decoding}



  \icmlsetsymbol{equal}{*}

    \begin{icmlauthorlist}
    \icmlauthor{Zhongyu Xiao}{yyy}
    \icmlauthor{Zhiwei Hao}{comp}
    \icmlauthor{Jianyuan Guo}{comp}
    \icmlauthor{Yong Luo}{sch}
    \icmlauthor{Jia Liu}{ose}
    \icmlauthor{Jie Xu}{loj}
    \icmlauthor{Han Hu}{yyy}
  \end{icmlauthorlist}

  \icmlaffiliation{yyy}{School of information and Electronics, Beijing Institute of Technology, Beijing}
  \icmlaffiliation{comp}{Department of Computer Science, City University of Hong Kong, Hong Kong}
  \icmlaffiliation{sch}{School of Computer Science, Wuhan University, Wuhan}
  \icmlaffiliation{ose}{School of Economics and Management, Communication University of China, Beijing}
  \icmlaffiliation{loj}{School of Science and Engineering, The Chinese University of Hong Kong (Shenzhen), Shenzhen}

  \icmlcorrespondingauthor{Han Hu}{hhu@bit.edu.cn}

  \icmlkeywords{Machine Learning, ICML}

  \vskip 0.3in
]



\printAffiliationsAndNotice{}  

\begin{abstract}
Diffusion Large Language Models (dLLMs) offer a compelling paradigm for natural language generation, leveraging parallel decoding and bidirectional attention to achieve superior global coherence compared to autoregressive models. While recent works have accelerated inference via KV cache reuse or heuristic decoding, they overlook the intrinsic inefficiencies within the block-wise diffusion process. Specifically, they suffer from spatial redundancy by modeling informative-sparse suffix regions uniformly and temporal inefficiency by applying fixed denoising schedules across all the decoding process. To address this, we propose Streaming-dLLM, a training-free framework that streamlines inference across both spatial and temporal dimensions. Spatially, we introduce attenuation guided suffix modeling to approximate the full context by pruning redundant mask tokens. Temporally, we employ a dynamic confidence aware strategy with an early exit mechanism, allowing the model to skip unnecessary iterations for converged tokens. Extensive experiments show that Streaming-dLLM achieves up to 68.2$\times$ speedup while maintaining generation quality, highlighting its effectiveness in diffusion decoding. The code is available at \href{https://github.com/xiaoshideta/Streaming-dLLM}{https://github.com/xiaoshideta/Streaming-dLLM}.
\end{abstract}

\section{Introduction}
\begin{figure}[ht]
	\begin{center}
		\centerline{\includegraphics[width=\columnwidth]{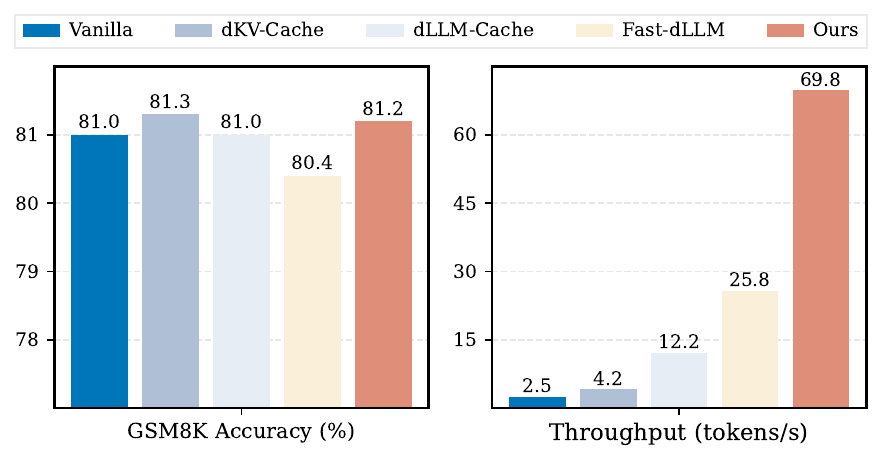}}
		\caption{
			Comparison of accuracy and throughput across different acceleration strategies.
			Our proposed method improves inference throughput while maintaining competitive accuracy compared to prior approaches.
		}
		\label{icml-historical}
	\end{center}
    \vspace{-0.5cm}
\end{figure}
Diffusion Large Language Models (dLLMs) have gained emerged as a promising generative paradigm in natural language processing~\cite{nie2025large,ye2025dream,zhu2025llada,zhu2025llada1,you2025llada,sahoo2024simple}. In contrast to traditional autoregressive Large Language models (LLMs)~\cite{achiam2023gpt,guo2025deepseek,dubey2024llama,yang2025qwen3,chen2025pangu}, dLLMs employ parallel decoding with bidirectional attention, enabling simultaneous information integration across all sequence positions and fostering superior global coherence and semantic consistency~\cite{li2022diffusion,arriola2025block}. This architectural shift is expected to alleviate the sequential inference bottleneck and the unidirectional context dependency inherent in autoregressive generation.

Despite these advantages, existing dLLMs lag significantly behind autoregressive models in inference speed~\cite{li2025survey, zhang2025survey}. This gap primarily arises from the intrinsic nature of diffusion decoding. In particular, token representations evolve continuously across denoising steps, which prevents the effective reuse of the highly optimized Key–Value (KV) Cache. Furthermore, parallel generation amplifies sensitivity to decoding uncertainty, often necessitating additional denoising iterations to preserve generation quality.

Consequently, improving the inference efficiency of dLLMs has attracted substantial attention recently~\cite{wu2023ar,wu2025fast1,liu2025wedlm,zhao2025d1,liu2025dllm, wei2025accelerating,chen2025dlm,tian2025next,ma2025dinfer}. For instance, dKV-Cache~\cite{ma2025dkv} exploits the partial stability of token states to enable limited cache reuse, while Fast-dLLM ~\cite{wu2025fast} adopts semi-autoregressive block-wise generation combined with confidence decoding. While these approaches achieve noticeable acceleration, they overlook the fundamental redundancy deeply rooted in the block-wise diffusion process. \textbf{Spatially}, mask tokens in the suffix region carry limited and homogeneous information, yet standard attention mechanisms process every suffix token at every step. Since the suffix region scales with the target generation length, the model wastes substantial computation on low value mask positions during the early and middle stages of diffusion. \textbf{Temporally}, the uncertainty within each generation block converges at varying rates. However, conventional thresholding strategies apply a static criterion, failing to adapt to this dynamic behavior. Consequently, high confidence tokens are often forced back into a masked state in subsequent iterations, which restricts the effective throughput of parallel decoding and may disrupt the refinement of truly uncertain tokens.

To address these challenges, we propose \textbf{Streaming-dLLM}, a training-free framework for fast and scalable inference. Our method is motivated by two key observations: the diminishing attention influence of distant suffix tokens and the dynamic evolution of token confidence within generation blocks. 
To tackle spatial inefficiency, we introduce attenuation-guided suffix modeling. Instead of attending to the entire suffix, we preserve only a small set of informative suffix blocks alongside positional cues of the final length to approximate the global context. This design drastically reduces the computational cost of iterative inference without compromising generation quality. To address temporal inefficiency, we propose a dynamic confidence aware parallel decoding strategy. This mechanism adaptively adjusts decoding thresholds based on both the diffusion stage and the intra-block confidence distribution. Furthermore, we incorporate an early exit mechanism to terminate decoding immediately upon predicting the End-of-Sequence (EOS) token, avoiding unnecessary computation for the remainder of the block sequence.
Streaming-dLLM integrates these complementary components to optimize decoding along both spatial and temporal dimensions. Extensive experiments demonstrate that our approach achieves up to 68.2$\times$ inference speedup without degrading downstream performance.

Our contributions are summarized as follows:

\begin{itemize}
  \item We propose attenuation guided suffix modeling, a training-free approach that accelerates inference by pruning redundant suffix attention during the decoding process.
  \item We develop a dynamic confidence aware parallel decoding strategy that flexibly adjusts parallelism according to real-time decoding uncertainty, enabling plug-and-play inference acceleration.
  \item Extensive experiments across multiple dLLMs and benchmarks demonstrate that Streaming-dLLM significantly improves inference throughput with negligible impact on generation quality. 
\end{itemize}

\section{Related works}
\subsection{Diffusion Language Models}

Diffusion models have emerged as a highly promising generative paradigm for text, serving as an alternative to autoregressive models based on a forward masking and reverse denoising process ~\cite{li2025survey, zhu2025latent}. In general, existing dLLMs are broadly categorized into continuous space ~\cite{li2022diffusion, yu2022latent, lin2023text, gao2024empowering, mahabadi2024tess} and discrete space ~\cite{austin2021structured, nie2025large, song2025seed, ye2025dream, zhu2025llada} approaches, depending on the space in which diffusion is performed.

Focusing on continuous space, Diffusion-LM ~\cite{li2022diffusion} pioneers this direction by applying Gaussian diffusion directly to token embeddings. GENIE ~\cite{lin2023text} introduces a large scale pretraining strategy with a paragraph level denoising objective that reconstructs corrupted text spans, enabling effective learning from massive corpora. In contrast, discrete space dLLMs define diffusion directly over the token vocabulary. D3PM \cite{austin2021structured} defines the forward noising procedure as a discrete Markov chain with a predefined transition matrix. LLaDA \cite{nie2025large} further advances this line of work by training discrete diffusion language models from scratch using a masked-token cross entropy objective. Overall, dLLMs have demonstrated strong empirical performance across multiple benchmarks and are becoming increasingly competitive with leading autoregressive language models.

\subsection{Acceleration Techniques for dLLMs}

To fully unleash the potential of dLLMs, recent studies have explored various inference~\cite{feng2025theoretical}. In terms of caching mechanisms, dKV-Cache ~\cite{ma2025dkv} introduces a delayed KV Cache to reduce redundant memory usage, while Elastic-Cache ~\cite{nguyen2025attention} leverages attention aware drift to perform adaptive KV refreshes, thereby improving cache efficiency. Regarding parallel decoding, Fast-dLLM ~\cite{wu2025fast} proposes a confidence threshold parallel decoding to accelerate the diffusion process. ReMDM ~\cite{wang2025remasking} introduces a remasking sampler at inference, allowing the model to revise previously generated tokens and achieve a flexible trade-off between computational cost and generation quality. For model pruning, DPad ~\cite{chen2025dpad} prunes redundant regions by sampling suffix windows, and Sparse-dLLM ~\cite{song2025sparse} dynamically removes low importance KV entries from both prefix and suffix tokens to reduce inference overhead. Nevertheless, these methods remain constrained by the intrinsic characteristics of the block-wise diffusion decoding, often exhibiting spatial redundancy and temporal inefficiency during generation, which limits their potential for further acceleration.

\section{Method}
\subsection{Preliminary}

Contrary to autoregressive LLMs, dLLMs perform inference through iterative decoding, where masked tokens are progressively reconstructed into the final output over discrete diffusion steps $T$. For a generation task with target length $L$, the initial sequence is defined as:
\begin{equation}
	x^{(0)}=[p_0,\underbrace{[\mathrm{MASK}],\ldots,[\mathrm{MASK}]}_L],
	\label{con:1}
\end{equation}
where $x^{(t)}$ denotes the state of the entire sequence at time $t\in[0, T]$, $p_0$ represents the input prompt and $[\mathrm{MASK}]$ indicates a masked token. The masked sequence of length $L$ is partitioned into $N$ non-overlapping blocks with $K$ tokens in each block, such that $L = N \times K$. For block ${B}_n^{(t)}, n\in\{0,\ldots,N-1\}$:
\begin{equation}
	\begin{aligned}
		{I}_n &= \{p_L+nK,\ldots,p_L+(n+1)K-1\}, \\
		{B}_n^{(t)} &= x_{{I}_n}^{(t)}
		= \left[ x_{p_L+nK}^{(t)}, \ldots, x_{p_L+(n+1)K-1}^{(t)} \right].
	\end{aligned}
	\label{con:2}
\end{equation}
Here, ${I}_n$ denotes the token indices in the $n$-th block, and $p_L$ is the index offset corresponding to the prompt length. This generation paradigm is referred to as block-wise diffusion decoding. In particular, when $K=1$, the decoding process degenerates into an autoregressive-like generation scheme. Contrastly, when $K=L$, it simplifies to pure diffusion decoding with all tokens generated in parallel.

Each block is decoded over $M$ diffusion steps ($T=N\times{M}$). At iteration $t\in\{0,\ldots,M-1\}$, the model predicts logits for all positions:
\begin{equation}
	z^{(t)}=f_\theta(x^{(t)}),
	\label{con:3}
\end{equation}
where $f_\theta$ denotes the dLLMs parameterized by $\theta$. Sampling and token prediction are then performed only for positions $i \in {B}_n$ where $x_i^{(t)} = [\mathrm{MASK}]$:
\begin{equation}
	\begin{aligned}
		c_i^{(t)} &= \max (\mathrm{Softmax}\!\left(z_i^{(t)}\right)), \\
		\hat{x}_i^{(t)} &= \arg\max (\mathrm{Softmax}\!\left(z_i^{(t)}\right)).
	\end{aligned}
	\label{con:4}
\end{equation}
Finally, token states are updated according to:
\begin{equation}
	x_i^{(t+1)}=
	\begin{cases}
		\hat{x}_i^{(t)}, & i \in \{i_k\mid S(\hat{x}_{i_k}^{(t)}, c_{i_k}^{(t)})\}, \\
		x_i^{(t)}, & \text{otherwise.}
	\end{cases}
\end{equation}
where $S(\cdot)$ denotes the token selection strategy, which determines the subset of positions to be updated based on predicted tokens and their corresponding confidence scores. The decoding iteration for block $n$ terminates when $\{i \in {B}_n \mid x_i = [\mathrm{MASK}]\} = \emptyset$, after which the model proceeds to decode block $n+1$.

\begin{figure}[ht]
	\begin{center}
		\centerline{\includegraphics[width=\columnwidth]{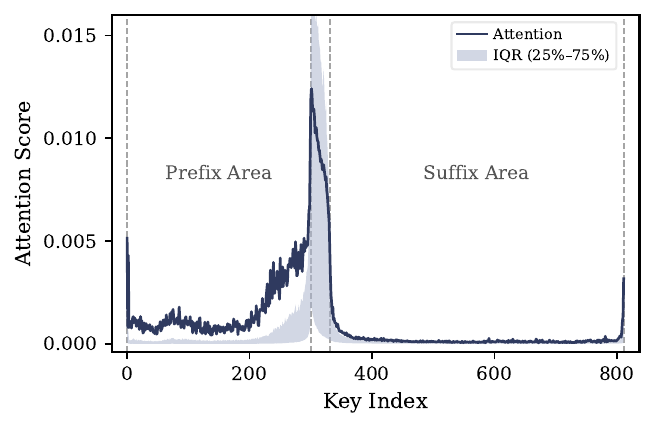}}
		\caption{
			Attention distribution at the final layer (Layer 31) of LLaDA-1.5, showing interactions between the current block and the full input sequence. The solid line denotes the mean attention score at each diffusion step, the shaded region represents the interquartile range (IQR, 25\%–75\%), and gray dashed lines delineate the prefix, current-generation and suffix regions. Attention is concentrated on a few neighboring suffix blocks and the final token and most intermediate suffix positions receive negligible attention, highlighting redundancy in the suffix during block generation.
		}
		\label{icml-fig2}
	\end{center}
    \vspace{-0.4cm}
\end{figure}
\subsection{Rethinking Inference of dLLMs}

Diffusion language models leverage bidirectional attention and parallel decoding to exploit global contextual information, thereby improving generation quality. However, whether this design remains necessary and efficient at every diffusion step during inference, especially for long text generation, has yet to be systematically examined.

In the diffusion decoding framework (Equation \ref{con:3}), bidirectional attention is computed over the full sequence $x^{(t)}$ at each prediction. Unlike autoregressive models, the input sequence of dLLMs consists of a semantically dense prompt region and a large number of ungenerated mask tokens. When the prompt is short and the generation length is large, the model is forced to repeatedly attend to an extensive masked region, which incurs substantial computational overhead. As masked tokens, these positions also provide a large amount of homogeneous semantic information, while whether such information is actually necessary or effective during generation remains unclear.

\begin{figure}[ht]
	\begin{center}
		\centerline{\includegraphics[width=\columnwidth]{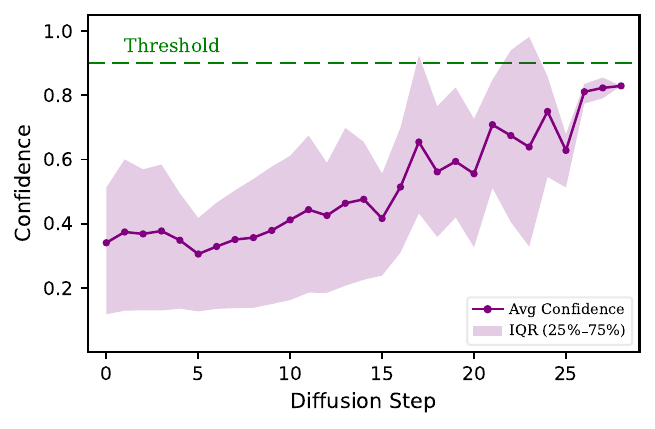}}
		\caption{
			Token confidence distribution during iterative block-wise generation of LLaDA-1.5 on GSM8K with a generation length of 256, showing that mean token confidence steadily increases over iterations. The solid line denotes the mean confidence at each diffusion step and the shaded area represents the interquartile range (IQR, 25\%--75\%). While a high confidence threshold ensures generation quality, it can be conservative, potentially delaying the acceptance of sufficiently confident tokens. Visualizations of additional blocks are provided in the ~\cref{sec:Details of Attention}.
		}
		\label{icml-fig3}
	\end{center}
    \vspace{-0.4cm}
\end{figure}

To investigate this, we analyze attention distribution between the currently generated block and the entire input sequence. As shown in~\cref{icml-fig2}, attention over the suffix region decays with distance. Although the suffix constitutes most of the sequence, many intermediate positions receive near-zero scores. Conversely, the current generation block primarily attends to a small number of neighboring blocks within the suffix region, as well as the token at the end position. This suggests that only a limited portion of the suffix mask region contains meaningful structural information, while the remaining positions largely serve as redundant placeholders. Further details and discussions regarding the~\cref{icml-fig2} are provided in the ~\cref{sec:Details of Attention}.

Meanwhile, we further examine the evolution behavior of token confidence during the parallel decoding process within the block. Previous method~\cite{wu2025fast} typically uses a fixed confidence threshold to determine whether a token should be updated, which implicitly assumes that the confidence distribution remains relatively stable during decoding. In light of this, we collect statistics from the iterative decoding process on 100 GSM8K samples and visualize the first generated block. As shown in~\cref{icml-fig3}, the number of remaining masked tokens within the current block steadily decreases as diffusion progresses, while the confidence distribution undergoes substantial contraction and shift. Moreover, we plot token confidences within the interquartile range (IQR, $25\%-75\%$), which better reflects the typical behavior of most tokens during decoding. The observation indicate that using a fixed confidence threshold is suboptimal: early steps require a high threshold to ensure generation quality, while later steps benefit from a lower threshold to allow timely token acceptance and improve parallel decoding efficiency.

\begin{figure*}[ht]
	\begin{center}
		\centerline{\includegraphics[width=\columnwidth*2]{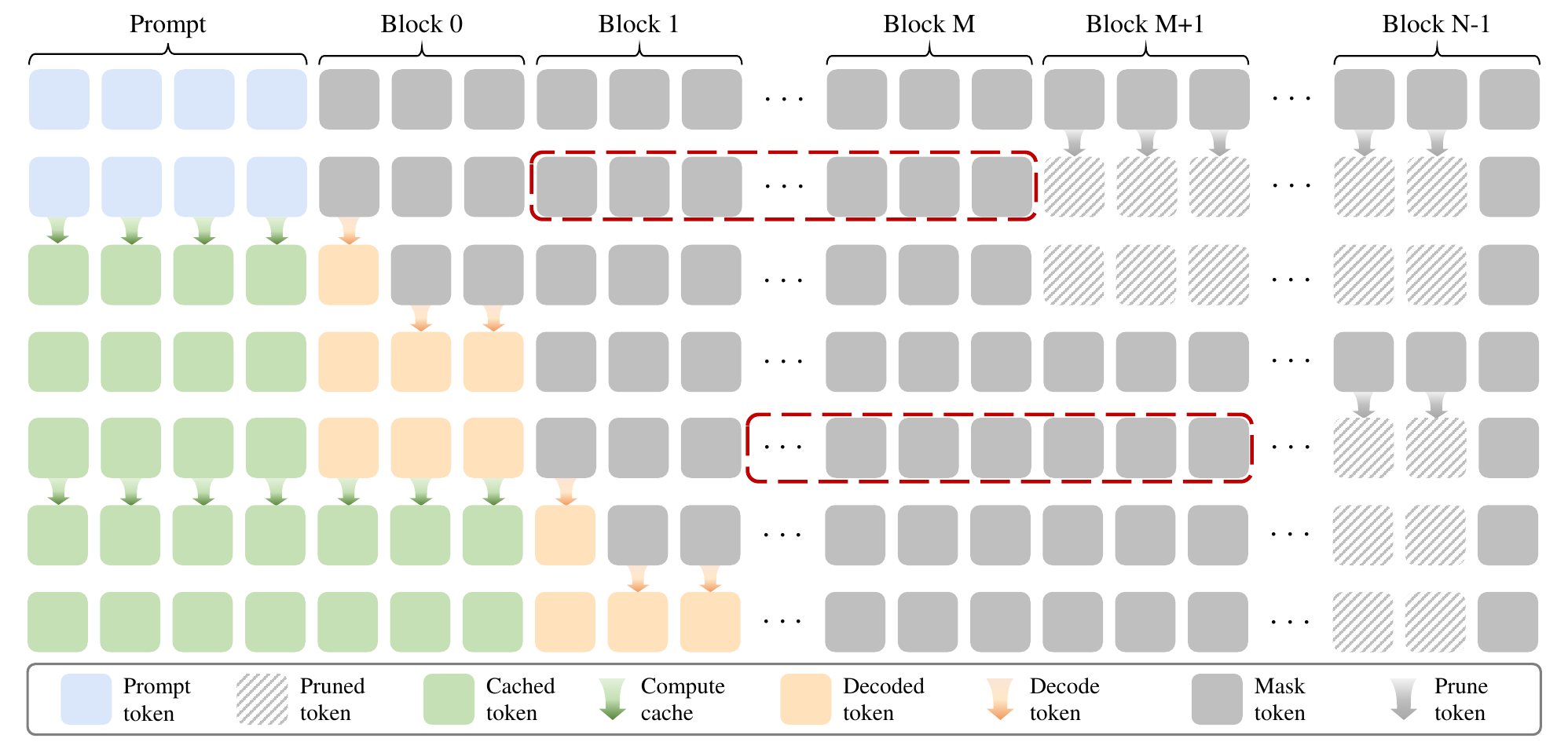}}
		\caption{
			Illustration of Attenuation-Guided Suffix Modeling. For each block, the nearest neighboring region following the current block is retained using a sliding window (red dashed box) and concatenated with the trailing position to form an approximate suffix region.
		}
		\label{icml-method}
	\end{center}
    \vspace{-0.4cm}
\end{figure*}

Overall, above analysis reveals two complementary inefficiencies in the reasoning process of dLLMs:
(1) Spatially, a substantial amount of computation is devoted to suffix mask tokens that carry limited semantic information, leading to a clear mismatch between computation and informative content. 
(2) Temporally, the fixed confidence decoding strategy fails to capture the dynamic evolution of uncertainty within a block, forcing tokens with sufficient confidence to participate in needless diffusion steps. These observations motivate our redesign of both the suffix modeling approach and the decoding strategy for the reasoning stage.

\subsection{Streaming-dLLM}

Motivated by the two inefficiencies identified above, we propose Streaming-dLLM, a training-free framework designed to accelerate dLLMs inference. The core insight is to streamline block-wise diffusion decoding by progressively eliminating redundant computation along both spatial and temporal dimensions. In response to spatial redundancy in the suffix region, when decoding the $n$-th block, Streaming-dLLM first applies attenuation guided suffix modeling to retain only a minimal set of structural cues that are most critical for generating the current block. In parallel, to reflect the dynamic confidence evolution, we further adapts the parallel decoding strategy during intra-block diffusion iterations according to the evolving token confidence distribution, allowing high confidence tokens to be finalized earlier. Additionally, we incorporate an early exit mechanism to terminate decoding once the EOS token is predicted.

\textbf{Attenuation Guided Suffix Modeling.} Following Equations~\ref{con:1} and~\ref{con:2}, the sequence at diffusion step $t$ is:
\begin{equation}
	x^{(t)}=[p_0,\hat{{B}}_0^{(t)},\ldots,\hat{{B}}_{c-1}^{(t)}, {B}_c^{(t)},\ldots,{B}_{N-1}^{(t)}], 
\end{equation}
where $\hat{{B}}_i^{(t)}, i \in [0, c-1]$ denotes the blocks that have already been decoded. Together with the input prompt $p_0$, they form the prefix region ${S}_{\text{prefix}}^{(t)}$. ${B}_c^{(t)}$ represents the block being decoded at diffusion step $t$, referred to as the current region ${S}_{\text{current}}^{(t)}$. The remaining blocks ${B}_j^{(t)}$ for $j \in (c, N-1]$ constitute the suffix region ${S}_{\text{suffix}}^{(t)}$.

\begin{table*}[ht]
	\centering
	\caption{
		Comparison of Dream-Base suite performance across four benchmarks at different generation lengths (256 and 512). Each cell reports accuracy (top) and decoding throughput with relative speedup over the Dream baseline (bottom; blue: tokens/s, orange: speedup). Results in bold indicate the highest score, while those underlined denote the second-best performance for each method. `\textsuperscript{*}' marks results we reproduced using the official implementation, as they were not reported in the original papers.}
	\renewcommand{\arraystretch}{1.16}
	\setlength{\tabcolsep}{6pt}
	\begin{center}
		\begin{small}
			\begin{tabular}{lcccccc}
				\toprule
				Benchmark & Gen Length & Dream & dKV-Cache & Prefix-Cache & Fast-dLLM & Ours  \\
				\midrule
				\multirow{4}{*}{\centering HumanEval (0-shot)}
				& 256 & 49.4 & 48.2$^{*}$ & \underline{53.7} & \textbf{54.3} & \cellcolor{mytan}\textbf{54.3} \\
				&       & \textcolor{blue}{20.4} (\textcolor{orange}{1$\times$})
				& \textcolor{blue}{21.5} (\textcolor{orange}{1.1$\times$})
				& \textcolor{blue}{32.0} (\textcolor{orange}{1.6$\times$})
				& \textcolor{blue}{53.7} (\textcolor{orange}{2.6$\times$})
				& \cellcolor{mytan}\textcolor{blue}{74.7} (\textcolor{orange}{3.7$\times$})\\
				& 512 & 54.3 & 49.4$^{*}$ & \textbf{54.9} & 54.3 & \cellcolor{mytan}\underline{54.6} \\
				&       & \textcolor{blue}{13.7} (\textcolor{orange}{1$\times$})
				& \textcolor{blue}{15.7} (\textcolor{orange}{1.1$\times$})
				& \textcolor{blue}{24.2} (\textcolor{orange}{1.8$\times$})
				& \textcolor{blue}{40.2} (\textcolor{orange}{2.9$\times$})
				& \cellcolor{mytan}\textcolor{blue}{72.3} (\textcolor{orange}{5.3$\times$})\\
				\midrule
				\multirow{4}{*}{GSM8K-CoT (5-shot)}
				& 256 & \textbf{74.8} & 73.6$^{*}$ & \underline{74.0} & 73.5 & \cellcolor{mytan}\underline{74.0} \\
				&       & \textcolor{blue}{9.0} (\textcolor{orange}{1$\times$}) 
				& \textcolor{blue}{17.0} (\textcolor{orange}{1.9$\times$})
				& \textcolor{blue}{31.5} (\textcolor{orange}{3.5$\times$})
				& \textcolor{blue}{47.9} (\textcolor{orange}{5.3$\times$}) 
				& \cellcolor{mytan}\textcolor{blue}{75.5} (\textcolor{orange}{8.4$\times$}) \\
				& 512 & \underline{74.2} & 71.6$^{*}$ & \underline{74.2} & 74.1 & \cellcolor{mytan}\textbf{74.7} \\
				&       & \textcolor{blue}{7.1} (\textcolor{orange}{1$\times$}) 
				& \textcolor{blue}{12.8} (\textcolor{orange}{1.8$\times$})
				& \textcolor{blue}{23.6} (\textcolor{orange}{3.3$\times$}) 
				& \textcolor{blue}{41.7} (\textcolor{orange}{5.9$\times$}) 
				& \cellcolor{mytan}\textcolor{blue}{94.1} (\textcolor{orange}{13.3$\times$}) \\
				\midrule
				\multirow{4}{*}{MBPP (3-shot)}
				& 256 & \textbf{56.6} & 54.0$^{*}$ & 53.2 & \underline{56.4} & \cellcolor{mytan}\underline{56.4} \\ 
				&       & \textcolor{blue}{11.0} (\textcolor{orange}{1$\times$})   
				& \textcolor{blue}{14.7} (\textcolor{orange}{1.3$\times$})
				& \textcolor{blue}{32.3} (\textcolor{orange}{2.9$\times$})
				& \textcolor{blue}{67.2} (\textcolor{orange}{6.1$\times$})
				& \cellcolor{mytan}\textcolor{blue}{80.2} (\textcolor{orange}{$7.3\times$})\\
				& 512 & \underline{55.6} & 53.0$^{*}$ & 53.8 & 55.2 & \cellcolor{mytan}\textbf{55.8}   \\
				&       & \textcolor{blue}{8.7} (\textcolor{orange}{1$\times$})
				& \textcolor{blue}{11.6} (\textcolor{orange}{1.3$\times$})
				& \textcolor{blue}{24.5} (\textcolor{orange}{2.8$\times$})
				& \textcolor{blue}{63.1} (\textcolor{orange}{7.3$\times$})
				& \cellcolor{mytan}\textcolor{blue}{92.4} (\textcolor{orange}{10.6$\times$})\\
				\midrule
				\multirow{4}{*}{\centering MATH (4-shot)}
				& 256 & \textbf{38.4} & 36.8$^{*}$ & 36.8 & \underline{37.6} & \cellcolor{mytan}\underline{37.6} \\
				&       & \textcolor{blue}{10.5} (\textcolor{orange}{1$\times$})   
				& \textcolor{blue}{14.6} (\textcolor{orange}{1.4$\times$})
				& \textcolor{blue}{32.5} (\textcolor{orange}{3.1$\times$})
				& \textcolor{blue}{62.6} (\textcolor{orange}{5.98$\times$})
				& \cellcolor{mytan}\textcolor{blue}{78.4} (\textcolor{orange}{7.5$\times$})\\
				& 512 & \textbf{39.8} & 38.5$^{*}$ & 38.0 & 39.3 & \cellcolor{mytan}\underline{39.4} \\
				&       & \textcolor{blue}{8.6} (\textcolor{orange}{1$\times$})
				& \textcolor{blue}{11.6} (\textcolor{orange}{1.3$\times$})
				& \textcolor{blue}{24.5} (\textcolor{orange}{2.8$\times$})
				& \textcolor{blue}{54.4} (\textcolor{orange}{6.3$\times$})
				& \cellcolor{mytan}\textcolor{blue}{96.0} (\textcolor{orange}{$11.2\times$})\\
				\bottomrule
				\label{tab:Dream-tps}
			\end{tabular}
		\end{small}
	\end{center}
    \vspace{-0.4cm}
\end{table*}

For the suffix region, we adopt a sliding window strategy that retains only $w$ blocks closest to ${S}_{\text{current}}^{(t)}$. In addition, we preserve the positional information of the final token to provide a coarse representation of the entire suffix while maintaining the logical ordering of tokens via RoPE~\cite{su2024roformer} position IDs.:
\begin{equation}
	\begin{gathered}
		\tilde{{I}}=\{{I}_c,\ldots,{I}_{c+w}\}\cup\{p_L+L\}, \\
		\tilde{{S}}_{\text{suffix}}^{(t)}=
		\bigcup_{{I}_{{S}} \in \tilde{{I}}}
		{S}_{\text{suffix}}^{(t)} .
	\end{gathered}
\end{equation}
The approximated suffix is then concatenated with the prefix and current regions to form the input sequence at diffusion step $t$, which is subsequently fed into the diffusion language model for prediction.
\begin{equation}
	\tilde{x}^{(t)}={S}_{\text{prefix}}^{(t)}\cup{S}_{\text{current}}^{(t)}\cup\tilde{{S}}_{\text{suffix}}^{(t)}, 
\end{equation}
This approximation is guided by the observation from ~\cref{icml-fig2}. During the generation of ${S}_{\text{current}}^{(t)}$, the suffix region primarily provides coarse structural constraints, such as positional continuity and termination cues, rather than detailed semantic information. Meanwhile, tokens in the distant suffix remain highly uncertain throughout most diffusion steps, limiting their contribution to refining the current block.

Accordingly, we retain only the suffix blocks adjacent to ${S}_{\text{current}}^{(t)}$ along with the terminal positional token, which preserves the essential structural signals required for stable decoding. This design significantly reduces redundant computation in both attention and denoising, while maintaining global coherence across blocks.

\textbf{Dynamic Confidence Aware Parallel Decoding.} During iterative block-wise decoding, the KV representations of the prefix region are computed once at the first diffusion step and reused across all subsequent steps within the same block. This approach significantly reduces redundant attention computation on the prefix while preserving information from previously decoded tokens ~\cite{wu2025fast}. 

In the prediction forward, the query vectors are constructed from the current generation region ${S}_{\text{current}}^{(t)}$ together with the approximated suffix region $\tilde{{S}}_{\text{suffix}}^{(t)}$. These queries attend to the KV representations of the entire sequence $\tilde{x}^{(t)}$ to perform attention computation. Attention scores are subsequently processed by the remaining model components to produce the prediction logits. Following Equation ~\ref{con:4}, we obtain generated token $\hat{x}_i^{(t)}$ for each position within query area, along with its corresponding confidence score $c_i^{(t)}$. The confidence scores are used to determine whether a token can be finalized at the current diffusion step. We define a confidence-aware selection function $S(\cdot)$ that decides whether a predicted token $\hat{x}_{i_k}^{(t)}$ is accepted according to its confidence $c_{i_k}^{(t)}$. Tokens whose confidence exceeds the threshold $\tau^{(t)}$ are finalized, while the remaining tokens stay masked and continue to be refined in subsequent diffusion iterations. When no token satisfies the confidence threshold, we finalize the masked token with the highest confidence. Formally, the selection rule can be defined as:
\begin{equation}
	S(\hat{x}_{i}^{(t)}, c_{i}^{(t)})=
	\begin{cases}
		\hat{x}_i^{(t)}, & c_{i}^{(t)}\ge\tau^{(t)}, \\
		\hat{x}_{\arg\max_{i} c_{i}^{(t)}}^{(t)}, & \forall i,\ c_{i}^{(t)} < \tau^{(t)}, \\
		\text{None}, & \text{otherwise.}
	\end{cases}
\end{equation}
To account for the evolving uncertainty during block-wise diffusion, we adopt an adaptive threshold that varies with the proportion of masked tokens in the current query. The threshold at step $t$ is defined as:
\begin{equation}
	\tau^{(t)}=\tau_0\left(1-\alpha(1-r_\mathrm{mask})\right),
\end{equation}
where $\tau_0$ denotes the base confidence threshold, $\alpha$ controls the adaptation strength, and $r_{\mathrm{mask}}$ represents the ratio of masked tokens at the current decoding step. By this adaptive design, we can further exploit the parallel decoding potential of diffusion language models, achieving a flexible trade-off between inference speed and accuracy.

\textbf{Early Exit For Block Diffusion.} To further improve inference efficiency, we introduce an early exit mechanism for block-wise diffusion decoding. During generation, if a block predicts the EOS token with high confidence, all subsequent blocks are skipped. This strategy prevents unnecessary computation on remaining blocks, reducing inference latency while ensuring that the generated sequence remains complete. By leveraging this property, Streaming-dLLM can adaptively reduce the decoding steps without compromising sequence integrity, further enhancing efficiency in text generation scenarios.

\section{Experiment}
\subsection{Experimental Setup}

To evaluate the effectiveness of Streaming-dLLM, we conduct experiments on representative dLLMs including Dream-v0-7B-Base~\cite{ye2025dream}, LLaDA-8B-Instruct~\cite{nie2025large}, and LLaDA-1.5~\cite{zhu2025llada}. The proposed method is compared with the corresponding vanilla implementations as well as recent state-of-the-art dLLM inference acceleration methods~\cite{ma2025dkv, wu2025fast}. We report results across a diverse set of domains, covering mathematical and scientific reasoning tasks (GSM8K~\cite{cobbe2021training}, MATH~\cite{hendrycks2020measuring}) as well as code generation benchmarks (HumanEval~\cite{chen2021evaluating}, MBPP~\cite{austin2021program}). In addition, we assess the models under varying generation lengths to examine the scalability and robustness of the proposed approach.

Our metrics capture both generation quality and inference efficiency, measured by benchmark accuracy, throughput (tokens per second, TPS) and inference latency per sample. For throughput computation, we count only non EOS tokens across the entire generated sequence, which ensures the reliability and comparability of the reported results. All experiments are conducted within the lm-eval framework to ensure consistency and reproducibility, using a single NVIDIA A800 80GB GPU. Additional optimization details are provided in the~\cref{sec:Optimization and Schedule}.

\begin{table*}[ht]
	\centering
	\caption{
		Comparison of LLaDA-1.5 suite performance across four benchmarks at different generation lengths (256 and 512). Each cell reports accuracy (top) and decoding throughput with relative speedup over the LLaDA-1.5 baseline (bottom; blue: tokens/s, orange: speedup). Results in bold indicate the highest score, while those underlined denote the second-best performance for each method. `\textsuperscript{*}' marks results we reproduced using the official implementation, as they were not reported in the original papers.}
	\renewcommand{\arraystretch}{1.16}
	\setlength{\tabcolsep}{6pt}
	\begin{center}
		\begin{small}
			\begin{tabular}{lcccccc}
				\toprule
				Benchmark & Gen Length & LLaDA-1.5 & dKV-Cache & Prefix-Cache & Fast-dLLM & Ours  \\
				\midrule
				\multirow{4}{*}{\centering HumanEval (0-shot)}
				& 256 & \textbf{43.9}$^{*}$ & \underline{40.2}$^{*}$ & 38.4$^{*}$ & 37.2$^{*}$ & \cellcolor{mytan}39.0 \\
				&       & \textcolor{blue}{6.4} (\textcolor{orange}{1$\times$})
				& \textcolor{blue}{6.6} (\textcolor{orange}{1.0$\times$})
				& \textcolor{blue}{10.9} (\textcolor{orange}{1.7$\times$})
				& \textcolor{blue}{19.1} (\textcolor{orange}{3.0$\times$})
				& \cellcolor{mytan}\textcolor{blue}{34.1} (\textcolor{orange}{5.3$\times$})\\
				& 512 & \textbf{40.5}$^{*}$ & \underline{40.2}$^{*}$ & 37.8$^{*}$ & 39.8$^{*}$ & \cellcolor{mytan}\underline{40.2} \\
				&       & \textcolor{blue}{2.9} (\textcolor{orange}{1$\times$})
				& \textcolor{blue}{3.3} (\textcolor{orange}{1.1$\times$})
				& \textcolor{blue}{4.8} (\textcolor{orange}{1.7$\times$})
				& \textcolor{blue}{13.6} (\textcolor{orange}{4.7$\times$})
				& \cellcolor{mytan}\textcolor{blue}{26.7} (\textcolor{orange}{9.2$\times$})\\
				\midrule
				\multirow{4}{*}{\centering GSM8K (5-shot)}
				& 256 & 80.5$^{*}$ & \underline{80.7}$^{*}$ & 80.6$^{*}$ & \underline{80.7} & \cellcolor{mytan}\textbf{80.8} \\
				&       & \textcolor{blue}{6.3} (\textcolor{orange}{1$\times$}) 
				& \textcolor{blue}{10.8} (\textcolor{orange}{1.7$\times$}) 
				& \textcolor{blue}{24.4} (\textcolor{orange}{3.9$\times$})
				& \textcolor{blue}{50.0} (\textcolor{orange}{7.9$\times$}) 
				& \cellcolor{mytan}\textcolor{blue}{66.2} (\textcolor{orange}{10.5$\times$}) \\
				& 512 & 81.0$^{*}$ & \textbf{81.3}$^{*}$ & 81.0$^{*}$ & 80.4 & \cellcolor{mytan}\underline{81.2} \\
				&       & \textcolor{blue}{2.5} (\textcolor{orange}{1$\times$}) 
				& \textcolor{blue}{4.2} (\textcolor{orange}{1.7$\times$}) 
				& \textcolor{blue}{8.2} (\textcolor{orange}{3.3$\times$})
				& \textcolor{blue}{25.8} (\textcolor{orange}{10.3$\times$}) 
				& \cellcolor{mytan}\textcolor{blue}{69.8} (\textcolor{orange}{28.0$\times$}) \\
				\midrule
				\multirow{4}{*}{\centering MBPP (3-shot)}
				& 256 & \underline{38.0}$^{*}$ & \textbf{38.2}$^{*}$ & 37.8$^{*}$ & 37.6$^{*}$ & \cellcolor{mytan}37.8 \\ 
				&       & \textcolor{blue}{2.2} (\textcolor{orange}{1$\times$})   
				& \textcolor{blue}{3.5} (\textcolor{orange}{1.6$\times$})
				& \textcolor{blue}{7.6} (\textcolor{orange}{3.5$\times$})
				& \textcolor{blue}{29.5} (\textcolor{orange}{13.4$\times$})
				& \cellcolor{mytan}\textcolor{blue}{54.7} (\textcolor{orange}{24.9$\times$})\\
				& 512 & \underline{38.2}$^{*}$ & 38.1$^{*}$ & 38.0$^{*}$ & 38.1$^{*}$ & \cellcolor{mytan}\textbf{38.4}   \\
				&       & \textcolor{blue}{0.9} (\textcolor{orange}{1$\times$})
				& \textcolor{blue}{1.5} (\textcolor{orange}{1.7$\times$})
				& \textcolor{blue}{2.8} (\textcolor{orange}{3.1$\times$})
				& \textcolor{blue}{16.5} (\textcolor{orange}{18.3$\times$})
				& \cellcolor{mytan}\textcolor{blue}{61.4} (\textcolor{orange}{68.2$\times$})\\
				\midrule
				\multirow{4}{*}{\centering MATH (4-shot)}
				& 256 & \underline{32.7}$^{*}$ & 31.8$^{*}$ & 32.5$^{*}$ & 32.6 & \cellcolor{mytan}\textbf{33.7} \\
				&       & \textcolor{blue}{7.8} (\textcolor{orange}{1$\times$})   
				& \textcolor{blue}{12.4} (\textcolor{orange}{1.6$\times$})
				& \textcolor{blue}{25.9} (\textcolor{orange}{3.3$\times$})
				& \textcolor{blue}{47.1} (\textcolor{orange}{6.0$\times$})
				& \cellcolor{mytan}\textcolor{blue}{66.2} (\textcolor{orange}{8.5$\times$})\\
				& 512 & \textbf{37.1}$^{*}$ & \underline{35.1}$^{*}$ & 35.0$^{*}$ & \underline{35.1} & \cellcolor{mytan}\underline{35.1} \\
				&       & \textcolor{blue}{4.8} (\textcolor{orange}{1$\times$})
				& \textcolor{blue}{7.5} (\textcolor{orange}{1.6$\times$})
				& \textcolor{blue}{13.9} (\textcolor{orange}{2.9$\times$})
				& \textcolor{blue}{38.3} (\textcolor{orange}{7.9$\times$})
				& \cellcolor{mytan}\textcolor{blue}{62.4} (\textcolor{orange}{13.0$\times$})\\
				\bottomrule
				\label{tab:LLaDA1.5-tps}
			\end{tabular}
		\end{small}
	\end{center}
    \vspace{-0.4cm}
\end{table*}

\subsection{Main Result}

As shown in~\cref{tab:Dream-tps}, Our method achieves 3.7$\times$–13.3$\times$ speedup across all benchmarks over the vanilla backbone. Compared with the state-of-the-art acceleration method, it provides 1.5$\times$–2.3$\times$ additional speedup on tasks with a generation length of 512. Meanwhile, our accuracy was comparable or slightly better, which demonstrates the effectiveness of our approach.

In addition, more experiments on LLaDA-1.5 show even greater throughput improvements, as presented in~\cref{tab:LLaDA1.5-tps}. In particular, Streaming-dLLM achieves a speedup of 68.2× on MBPP tasks with a generation length of 512, which substantially exceeds that of Fast-dLLM~\cite{wu2025fast}. Similar performance improvements are observed on the LLaDA-Instruct and more results are provided in~\cref{sec:More Experiments}.

Additional results on inference latency are provided in the ~\cref{sec:Inference Latency}. Compared with Fast-dLLM, Streaming-dLLM achieves up to 85.5\% reduction in inference latency per sample, indicating that our method not only improves throughput but also shortens real world response time. Overall, these results highlight the effectiveness and generalizability of our method across multiple model and benchmarks.

\subsection{Ablation Result}

In order to comprehensively evaluate our approach, we conduct ablation study on the GSM8K dataset using LLaDA-1.5 as the default model. 

\begin{table}[t]
	\caption{Ablation study on the components of our proposed method. The results are on the GSM8K dataset (Dream uses the GSM8K-CoT version) using LLaDA-1.5 for a generation length of 512. Each component contributes to the overall performance. \textbf{Suf.}: Attenuation Guided Suffix Modeling, \textbf{Dyn.}: Dynamic Confidence Aware Parallel Decoding, \textbf{Exit.}: Early Exit for Block Diffusion.}
	\label{tab:ablation}
	\begin{center}
		\begin{small}
			\begin{tabular}{c|ccc|cc}
				\toprule
				\textbf{Model} & \textbf{Suf.} & \textbf{Dyn.} & \textbf{Exit.} & \textbf{Acc.(\%)} & \textbf{Th.(tok/s)} \\
				\midrule
				\multirow{4}{*}{Dream} & {$\times$} & {$\times$} & {$\times$} & 74.1 & 41.7   \\
				& {$\checkmark$} & {$\times$} & {$\times$} & 74.6  & 76.8   \\
				& {$\checkmark$} & {$\checkmark$} & {$\times$} & 74.7 & 82.2    \\
				& {$\checkmark$} & {$\checkmark$} & {$\checkmark$} & \textbf{74.7} & \textbf{94.1}  \\
				\midrule
				\multirow{4}{*}{LLaDA} & {$\times$} & {$\times$} & {$\times$} & 77.2 & 32.1    \\
				& {$\checkmark$} & {$\times$} & {$\times$} & 78.6 & 46.9  \\
				& {$\checkmark$} & {$\checkmark$} & {$\times$} & 78.7 & 52.4   \\
				& {$\checkmark$} & {$\checkmark$} & {$\checkmark$} & \textbf{78.7} & \textbf{65.9}   \\
				\midrule
				\multirow{4}{*}{LLaDA-1.5} & {$\times$} & {$\times$} & {$\times$} & 80.4 & 25.8    \\
				& {$\checkmark$} & {$\times$} & {$\times$} & \textbf{81.4} & 44.9   \\
				& {$\checkmark$} & {$\checkmark$} & {$\times$} & 81.2 & 50.3   \\
				& {$\checkmark$} & {$\checkmark$} & {$\checkmark$} & 81.2 & \textbf{69.8}   \\
				\bottomrule
			\end{tabular}
		\end{small}
	\end{center}
	\vspace{-0.4cm}
\end{table}

\textbf{Ablation of each module.} To evaluate the effectiveness of the key components in our approach, we conduct ablation studies on three core modules: Attenuation Guided Suffix Modeling, Dynamic Confidence Aware Parallel Decoding and Early Exit for Block Diffusion. The results are summarized in ~\cref{tab:ablation}. When all three modules are enabled, the model achieves the largest throughput improvement, and this trend remains consistent across different dLLMs. This improvement is partly attributable to attenuation guided suffix modeling, which reduces computational complexity and may also enhance generation quality by mitigating interference from redundant suffix regions, as evidenced by a slight yet consistent gain. Such findings highlight its potential as a direction for future study in general dLLMs. Then, when dynamic confidence aware parallel decoding is activated, the generation quality exhibits moderate fluctuations. We attribute this behavior to the asynchronous evolution of token level confidence under parallel decoding, which introduces mild uncertainty perturbations at certain diffusion iterations. Finally, introducing early exit does not degrade generation quality, suggesting that the original diffusion language model is inherently robust during inference and that terminating diffusion iterations early under appropriate confidence conditions is both feasible and safe. From these experiments, jointly enabling the three proposed modules yields the strongest overall results, demonstrating that each component in our framework is essential for achieving improved performance.

\begin{table}[t]
	\caption{Impact of few-shots on Accuracy and Speedup for LLaDA-1.5 on GSM8K (5-Shot).}
	\label{tab:ablation-1}
	\begin{center}
		\begin{small}
			\begin{tabular}{l|ccc}
				\toprule
				\textbf{Model} & \textbf{3-shot} & \textbf{5-shot} & \textbf{8-shot}  \\
				\midrule
				\multirow{2}{*}{\centering LLaDA-1.5}  & 80.1 & 81.0 & 81.7    \\ 
				& \textcolor{blue}{3.4} (\textcolor{orange}{1$\times$})
				& \textcolor{blue}{2.5} (\textcolor{orange}{1$\times$}) 
				& \textcolor{blue}{1.7} (\textcolor{orange}{1$\times$}) \\
                \midrule
				\multirow{2}{*}{\centering Fast-dLLM}  & 79.7 & 80.4 & 79.9    \\
				& \textcolor{blue}{36.5} (\textcolor{orange}{10.7$\times$})
				& \textcolor{blue}{25.8} (\textcolor{orange}{10.3$\times$}) 
				& \textcolor{blue}{20.9} (\textcolor{orange}{12.3$\times$}) \\
                \midrule
				\multirow{2}{*}{\centering Ours}  &80.2  & 81.2 & 81.5     \\
				& \textcolor{blue}{77.4} (\textcolor{orange}{22.8$\times$})
				& \textcolor{blue}{69.8} (\textcolor{orange}{28.0$\times$}) 
				& \textcolor{blue}{61.3} (\textcolor{orange}{36.1$\times$}) \\
				\bottomrule
			\end{tabular}
		\end{small}
	\end{center}
	\vspace{-0.2cm}
\end{table}

\begin{table}[t]
	\caption{Impact of generation length on Accuracy and Speedup for LLaDA-1.5 on GSM8K (5-Shot).}
	\label{tab:ablation-2}
	\begin{center}
		\begin{small}
			\begin{tabular}{l|ccc}
				\toprule
				\textbf{Model} & \textbf{512} & \textbf{1024} & \textbf{2048}  \\
				\midrule
				\multirow{2}{*}{\centering LLaDA-1.5}  & 81.0 & 81.2 & 81.2    \\ 
				& \textcolor{blue}{2.5} (\textcolor{orange}{1$\times$})
				& \textcolor{blue}{0.8} (\textcolor{orange}{1$\times$}) 
				& \textcolor{blue}{0.3} (\textcolor{orange}{1$\times$}) \\
                \midrule
				\multirow{2}{*}{\centering Fast-dLLM}  & 80.4 & 80.5 & 80.7    \\
				& \textcolor{blue}{25.8} (\textcolor{orange}{10.3$\times$})
				& \textcolor{blue}{12.3} (\textcolor{orange}{15.4$\times$}) 
				& \textcolor{blue}{5.1} (\textcolor{orange}{17.0$\times$}) \\
                \midrule
				\multirow{2}{*}{\centering Ours}  & 81.2 & 81.2 & 82.0     \\
				& \textcolor{blue}{69.8} (\textcolor{orange}{28.0$\times$})
				& \textcolor{blue}{64.9} (\textcolor{orange}{81.1$\times$}) 
				& \textcolor{blue}{67.6} (\textcolor{orange}{225.3$\times$}) \\
				\bottomrule
			\end{tabular}
		\end{small}
	\end{center}
	\vspace{-0.4cm}
\end{table}

\textbf{Prefill and generation length.} We further investigate the effect of prefill and generation length. As shown in ~\cref{tab:ablation-1}, increasing the prefill length from 3-shot to 8-shot reduces throughput for both Fast-dLLM and Streaming-dLLM. Nevertheless, compared to Fast-dLLM, the speedup of our method increases from 2.1$\times$ to 2.9$\times$, indicating that the proposed mechanism remains effective at alleviating inference time computational overhead even under longer prefill settings. On the other hand, as reported in ~\cref{tab:ablation-2}, our method exhibits a pronounced advantage over Fast-dLLM in long-context generation. 
\begin{table}[ht]
	\caption{Impact of trailing positional information.}
	\begin{center}
		\begin{small}
			\begin{tabular}{l|cc}
				\toprule
				\textbf{Model} & \textbf{Trailing Position} & \textbf{Accuracy (\%)} \\
				\midrule
				\multirow{2}{*}{Dream}     & $\times$   & 72.8  \\
				& $\checkmark$  & \textbf{74.7} \\
				\midrule
				\multirow{2}{*}{LLaDA}     & $\times$  &  77.5   \\
				& $\checkmark$  &  \textbf{78.7}   \\
				\midrule
				\multirow{2}{*}{LLaDA-1.5} & $\times$  &  79.6   \\
				& $\checkmark$  &  \textbf{81.2}   \\
				\bottomrule
			\end{tabular}
		\end{small}
	\end{center}
	\vspace{-0.2cm}
	\label{tab:ablation-3}
\end{table}
When the generation length reaches 2048 tokens, it achieves a speedup of up to 225.3× relative to the vanilla model. Moreover, across all evaluated settings, our approach consistently outperforms Fast-dLLM in terms of generation quality, demonstrating that our acceleration strategy not only improves efficiency but also enables dLLMs to better adapt to block-wise diffusion inference.

\textbf{Trailing positional information.} Discussion on the necessity of retaining trailing positional information is summarized in ~\cref{tab:ablation-3}. It can be observed that omitting the trailing positional information leads to a noticeable drop in generation accuracy. Remarkably, this outcome is both intuitive and theoretically justified. When trailing positional information is not preserved, the model theoretical performance depends mainly on the block size and the number of retained suffix blocks, rather than the overall generation length. In practice, existing dLLMs typically rely on entire sequence positional information for global modeling, yet the experimental results suggest that this is not strictly required, as a small portion of the suffix combined with the final positional information can approximate the overall behavior. This finding provides new insight into enabling dLLMs to strike a more suitable balance between autoregressive and pure diffusion generation paradigms.

\begin{figure}[ht]
	\begin{center}
		\centerline{\includegraphics[width=\columnwidth]{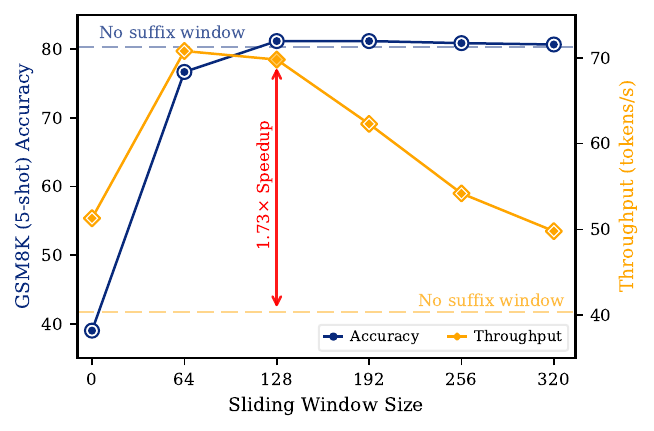}}
		\caption{
			Ablation study on the sliding window size (no suffix windows, mean $size=512$).
		}
		\label{tab:ablation-4}
	\end{center}
    \vspace{-0.4cm}
\end{figure}
\begin{figure}[ht]
	\begin{center}
		\centerline{\includegraphics[width=\columnwidth]{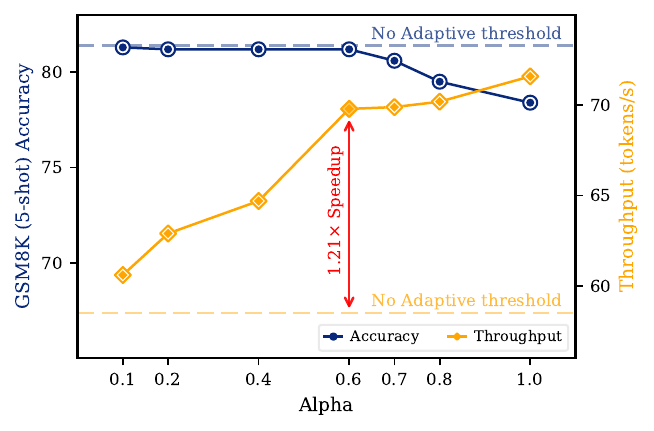}}
		\caption{
			Effect of the parallel decoding parameter $\alpha$ (no adaptive threshold, mean $\alpha=0$).
		}
		\label{tab:ablation-5}
	\end{center}
    \vspace{-0.4cm}
\end{figure}

\textbf{Hyperparameters.} We further conduct ablation studies on key hyperparameters. We first examine the effect of the sliding window size $w$ in Attenuation Guided Suffix Modeling. This parameter controls the fraction of the suffix region retained, balancing generation quality and inference efficiency. As shown in ~\cref{tab:ablation-4}, both accuracy and throughput improve as the window size increases. Once the window exceeds a certain threshold, accuracy saturates while throughput begins to decline. When $w=128$, a favorable balance is achieved, where accuracy is slightly improved and a 1.73× speedup is obtained relative to the full window size of 512. This result indicates that most suffix tokens contribute marginally to the generation of the current block, whose information can be sufficiently captured by a more compact near area, thereby confirming the presence of substantial computational redundancy in the suffix region.

We next study the effect of the parameter $\alpha$ , which controls parallel decoding. As reported in ~\cref{tab:ablation-5}, increasing $\alpha$  from 0.1 to 0.6 gradually improves throughput. However, further increasing $\alpha$  results in a degradation of overall performance, suggesting that overly aggressive threshold adjustments can adversely affect decoding stability and generation quality. This is likely due to premature parallel decoding of tokens that have not fully converged, , causing confidence misalignment that accumulates across subsequent iterations.

\subsection{Extension to Block-Causal dLLMs}

We further examined whether the redundancy reduction principle of Streaming-dLLM can be applied to block-causal dLLMs. As a representative instance, we evaluated Open Pangu 7B Base~\cite{tian2025next}, which follows a context-causal next-block decoding topology. Unlike Dream and LLaDA, where future masked suffix tokens are explicitly visible and therefore need to be pruned by attenuation-guided suffix modeling, block-causal models already remove the distant suffix through their causal decoding topology. In this setting, the spatial redundancy reduction module degenerates into a topology-aware special case, while the temporal redundancy reduction module can still be applied to improve token commit decisions. We therefore instantiate the dynamic decoding idea of Streaming-dLLM with stability-aware commit signals, without modifying model parameters or retraining the backbone.

\begin{table}[t]
	\caption{Representative results of applying the temporal decoding extension to Open Pangu 7B Base. Each cell reports accuracy (top) and decoding throughput with relative speedup over the baseline (bottom; blue: tokens/s, orange: speedup).}
	\label{tab:pangu-extension}
	\begin{center}
		\begin{small}
			\begin{tabular}{lcc}
				\toprule
				\textbf{Benchmark} & \textbf{Open Pangu} & \textbf{Ours} \\
				\midrule
				\multirow{2}{*}{\centering GSM8K}
				& 69.29 & \cellcolor{mytan}\textbf{75.82} \\
				& \textcolor{blue}{11.8} (\textcolor{orange}{1$\times$})
				& \cellcolor{mytan}\textcolor{blue}{18.3} (\textcolor{orange}{1.6$\times$}) \\
				\midrule
				\multirow{2}{*}{\centering MATH}
				& 41.14 & \cellcolor{mytan}\textbf{41.46} \\
				& \textcolor{blue}{9.7} (\textcolor{orange}{1$\times$})
				& \cellcolor{mytan}\textcolor{blue}{13.1} (\textcolor{orange}{1.4$\times$}) \\
				\midrule
				\multirow{2}{*}{\centering HumanEval}
				& 47.56 & \cellcolor{mytan}\textbf{48.17} \\
				& \textcolor{blue}{10.4} (\textcolor{orange}{1$\times$})
				& \cellcolor{mytan}\textcolor{blue}{14.6} (\textcolor{orange}{1.4$\times$}) \\
				\midrule
				\multirow{2}{*}{\centering MMLU-Pro}
				& 51.65 & \cellcolor{mytan}\textbf{51.65} \\
				& \textcolor{blue}{16.6} (\textcolor{orange}{1$\times$})
				& \cellcolor{mytan}\textcolor{blue}{25.4} (\textcolor{orange}{1.5$\times$}) \\
				\midrule
				\multirow{2}{*}{\centering BBH}
				& 51.33 & \cellcolor{mytan}\textbf{51.66} \\
				& \textcolor{blue}{13.1} (\textcolor{orange}{1$\times$})
				& \cellcolor{mytan}\textcolor{blue}{20.1} (\textcolor{orange}{1.5$\times$}) \\
				\midrule
				\multirow{2}{*}{\centering CMMLU}
				& \textbf{75.46} & \cellcolor{mytan}74.72 \\
				& \textcolor{blue}{18.2} (\textcolor{orange}{1$\times$})
				& \cellcolor{mytan}\textcolor{blue}{27.9} (\textcolor{orange}{1.5$\times$}) \\
				\bottomrule
			\end{tabular}
		\end{small}
	\end{center}
	\vspace{-0.2cm}
\end{table}

As shown in~\cref{tab:pangu-extension}, the adapted temporal decoding strategy improves or maintains accuracy on five of the six benchmarks and increases throughput by 1.4$\times$-1.6$\times$ across all evaluated tasks, with the largest accuracy gain on GSM8K. These results indicate that Streaming-dLLM is not limited to fully bidirectional dLLMs: for context-causal next-block models, the spatial component becomes implicit, while the temporal component remains an effective plug-in decoding strategy.

\section{Conclusion}
We presented Streaming-dLLM, an efficient inference acceleration framework for dLLMs. The proposed method introduces Attenuation Guided Suffix Modeling to approximate the suffix region of the sequence by retaining only a small set of structurally informative regions. This design substantially reduces redundant computation in the spatial dimension during block-wise decoding. Building upon this, we further propose dynamic confidence-aware parallel decoding, which adaptively adjusts the degree of parallel decoding according to the dynamic evolution of token confidence across diffusion iterations. Combined with an early-exit mechanism, this strategy effectively avoids unnecessary denoising iterations for tokens that have already achieved sufficient confidence, thereby further reducing inference overhead in the temporal dimension. Overall, Streaming-dLLM effectively addresses the dual inefficiencies of existing dLLMs during inference in both the spatial and temporal dimensions. Extensive evaluations across multiple dLLMs benchmarks show that our method preserves Comparable performance while achieving up to 68.3× inference speedup, demonstrating strong practicality and scalability.

\section*{Impact Statement}
This paper presents work whose goal is to advance the field of machine learning by improving the efficiency of dLLMs. While more efficient generative models may broaden access and reduce energy consumption, they could also lower barriers to misuse, which is a general concern in this area of research. Aside from these widely recognized implications, we do not believe our work raises additional ethical or societal issues that must be specifically highlighted here.

\bibliography{example_paper}
\bibliographystyle{icml2026}

\newpage
\appendix
\onecolumn
\section{Details of Attention and Token Confidence Sampling Procedure}
\label{sec:Details of Attention}

For~\cref{icml-fig2}, we analyze the attention distribution at the final layer (Layer 31) of LLaDA-1.5, which captures interactions between the currently generated block and the full input sequence. Statistics are collected from 300 GSM8K samples with a generation length of 512. For each sample, the first 300 tokens of the prompt are treated as the prefix area; tokens in [300:300+block\_{size}] correspond to the currently generated block, while tokens in [300+block\_{size}:] form the suffix area. In the analysis, the solid line represents the mean attention score at each diffusion step, and the shaded area indicates the interquartile range (25\%–75\%). The results indicate that attention is largely concentrated on a few neighboring suffix blocks and the final token, whereas most intermediate positions in the suffix area receive negligible attention. This pattern suggests substantial redundancy in the suffix region during block-wise generation.

For~\cref{icml-fig3}, we further examine the token confidence distribution during iterative block-wise generation of LLaDA-1.5 on GSM8K with a generation length of 256. Statistics are collected from the iterative decoding process of 100 samples, focusing on the first generated block. The results show that the mean token confidence steadily increases as iterations progress. While applying a high confidence threshold helps ensure generation quality, it can also be conservative, potentially delaying the acceptance of tokens that are already sufficiently confident. 

\begin{figure}[ht]
	\centering
	\begin{minipage}[b]{0.48\textwidth}
		\centering
		\includegraphics[width=\textwidth]{fig/fig3}
		\caption{Token confidence distribution of block 0.}
        \label{fig:co0}
	\end{minipage}
	\hfill
	\begin{minipage}[b]{0.48\textwidth}
		\centering
		\includegraphics[width=\textwidth]{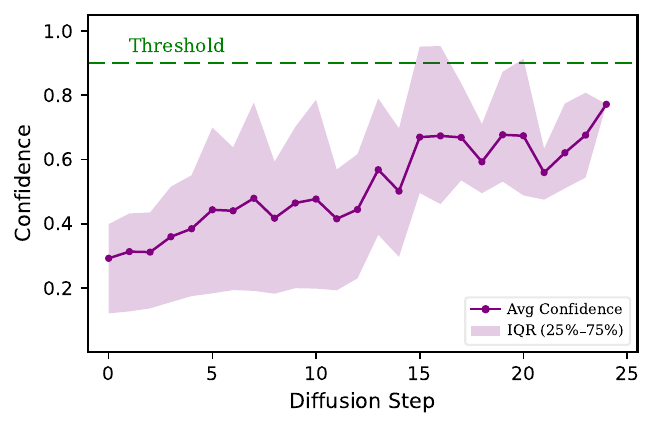}
		\caption{Token confidence distribution of block 1.}
		\label{fig:co1}
	\end{minipage}
	\vspace{-0.4cm}
\end{figure}

\begin{figure}[ht]
	\centering
	\begin{minipage}[b]{0.48\textwidth}
		\centering
		\includegraphics[width=\textwidth]{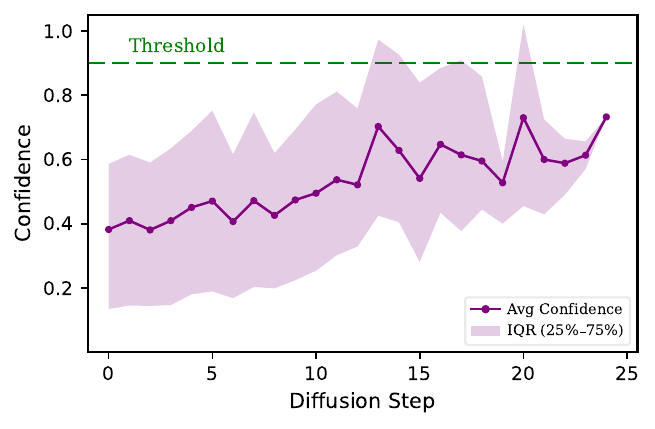}
		\caption{Token confidence distribution of block 2.}
		\label{fig:co2}
	\end{minipage}
	\hfill
	\begin{minipage}[b]{0.48\textwidth}
		\centering
		\includegraphics[width=\textwidth]{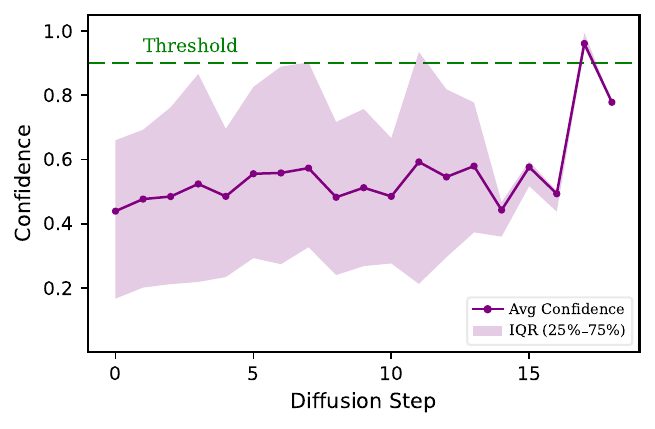}
		\caption{Token confidence distribution of block 3.}
		\label{fig:co3}
	\end{minipage}
	\vspace{-0.4cm}
\end{figure}

\begin{figure}[ht]
	\centering
	\begin{minipage}[b]{0.48\textwidth}
		\centering
		\includegraphics[width=\textwidth]{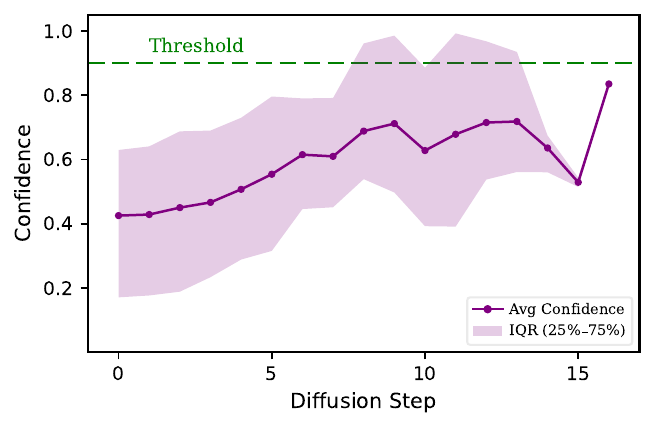}
		\caption{Token confidence distribution of block 4.}
		\label{fig:co4}
	\end{minipage}
	\hfill
	\begin{minipage}[b]{0.48\textwidth}
		\centering
		\includegraphics[width=\textwidth]{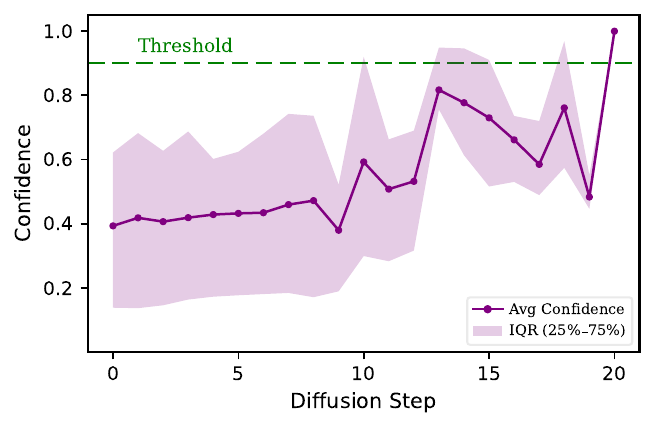}
		\caption{Token confidence distribution of block 5.}
		\label{fig:co5}
	\end{minipage}
\end{figure}

\begin{figure}[ht]
	\centering
	\begin{minipage}[b]{0.48\textwidth}
		\centering
		\includegraphics[width=\textwidth]{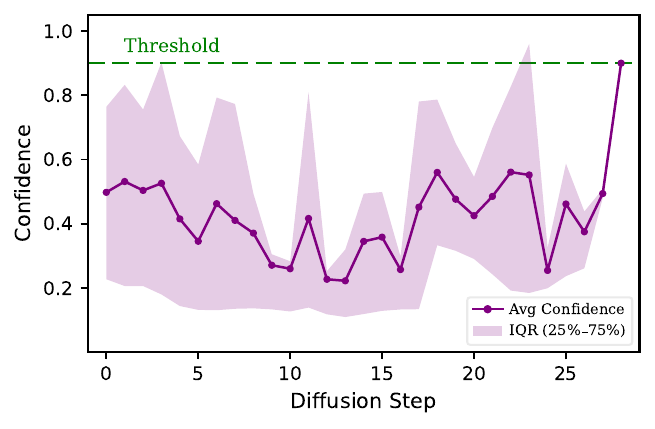}
		\caption{Token confidence distribution of block 6.}
		\label{fig:co6}
	\end{minipage}
	\hfill
	\begin{minipage}[b]{0.48\textwidth}
		\centering
		\includegraphics[width=\textwidth]{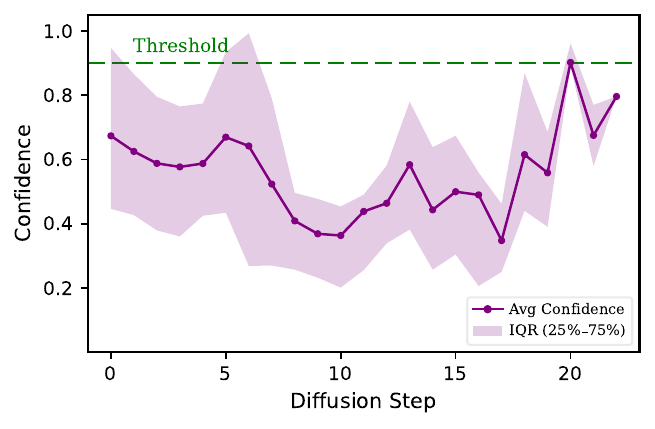}
		\caption{Token confidence distribution of block 7.}
		\label{fig:co7}
	\end{minipage}
\end{figure}

Additional visualizations for more generation blocks are shown in the ~\cref{fig:co0,fig:co1,fig:co2,fig:co3,fig:co4,fig:co5,fig:co6,fig:co7}. From these results, we observe that: (1) As iterations progress, blocks generated later tend to start with higher initial mean confidence, resulting in faster decoding. (2) In some decoding processes, the mean token confidence temporarily decreases. We attribute this behavior to a high confidence threshold, which forces many tokens back into the masked state, thereby affecting subsequent predictions. These observations further highlight the limitations of using a fixed confidence threshold in iterative block-wise generation.

\begin{table*}[ht]
	\centering
	\caption{
		Comparison of LLaDA-Instruct suite performance across four benchmarks at different generation lengths (256 and 512). Each cell reports accuracy (top) and decoding throughput with relative speedup over the LLaDA baseline (bottom; blue: tokens/s, orange: speedup). Results in bold indicate the highest score, while those underlined denote the second-best performance for each method. `\textsuperscript{*}' marks results we reproduced using the official implementation, as they were not reported in the original papers.
	}
	\renewcommand{\arraystretch}{1.16}
	\setlength{\tabcolsep}{6pt}
	\begin{center}
		\begin{small}
			\begin{tabular}{lcccccc}
				\toprule
				Benchmark & Gen Length & LLaDA & dKV-Cache & Prefix-Cache & Fast-dLLM & Ours  \\
				\midrule
				\multirow{4}{*}{\centering HumanEval (0-shot)}
				& 256 & 41.5 & 39.6$^{*}$ & 42.7 & \underline{43.3} & \cellcolor{mytan}\textbf{43.4} \\
				&       & \textcolor{blue}{18.4} (\textcolor{orange}{1$\times$})
				& \textcolor{blue}{18.9} (\textcolor{orange}{1.0$\times$})
				& \textcolor{blue}{30.6} (\textcolor{orange}{1.7$\times$})
				& \textcolor{blue}{71.5} (\textcolor{orange}{3.9$\times$})
				& \cellcolor{mytan}\textcolor{blue}{93.3} (\textcolor{orange}{5.1$\times$})\\
				& 512 & 43.9 & \underline{45.1}$^{*}$ & \textbf{45.7} & 44.5 & \cellcolor{mytan}44.5 \\
				&       & \textcolor{blue}{11.9} (\textcolor{orange}{1$\times$})
				& \textcolor{blue}{14.3} (\textcolor{orange}{1.2$\times$})
				& \textcolor{blue}{21.8} (\textcolor{orange}{1.8$\times$})
				& \textcolor{blue}{52.0} (\textcolor{orange}{4.4$\times$})
				& \cellcolor{mytan}\textcolor{blue}{75.8} (\textcolor{orange}{6.4$\times$})\\
				\midrule
				\multirow{4}{*}{\centering GSM8K (5-shot)}
				& 256 & \underline{79.3} & 76.5$^{*}$ & \textbf{79.5} & 78.5 & \cellcolor{mytan}78.5 \\
				&       & \textcolor{blue}{6.6} (\textcolor{orange}{1$\times$}) 
				& \textcolor{blue}{11.6} (\textcolor{orange}{1.8$\times$})
				& \textcolor{blue}{26.0} (\textcolor{orange}{3.9$\times$})
				& \textcolor{blue}{50.9} (\textcolor{orange}{7.7$\times$}) 
				& \cellcolor{mytan}\textcolor{blue}{66.2} (\textcolor{orange}{10.0$\times$}) \\
				& 512 & 77.5 & \underline{77.6}$^{*}$ & 77.0 & 77.2 & \cellcolor{mytan}\textbf{78.7} \\
				&       & \textcolor{blue}{3.2} (\textcolor{orange}{1$\times$}) 
				& \textcolor{blue}{5.5} (\textcolor{orange}{1.7$\times$})
				& \textcolor{blue}{11.0} (\textcolor{orange}{3.4$\times$})
				& \textcolor{blue}{32.1} (\textcolor{orange}{10.0$\times$}) 
				& \cellcolor{mytan}\textcolor{blue}{65.9} (\textcolor{orange}{20.6$\times$}) \\
				\midrule
				\multirow{4}{*}{\centering MBPP (3-shot)}
				& 256 & 29.4 & 26.2$^{*}$ & \underline{29.6} & 28.2 & \cellcolor{mytan}\textbf{42.0} \\ 
				&       & \textcolor{blue}{5.9} (\textcolor{orange}{1$\times$})   
				& \textcolor{blue}{10.0} (\textcolor{orange}{1.7$\times$})
				& \textcolor{blue}{19.7} (\textcolor{orange}{3.3$\times$})
				& \textcolor{blue}{44.3} (\textcolor{orange}{7.5$\times$})
				& \cellcolor{mytan}\textcolor{blue}{64.2} (\textcolor{orange}{10.9$\times$})\\
				& 512 & \underline{14.8} & 14.6$^{*}$ & 13.4 & 13.8 & \cellcolor{mytan}\textbf{40.8}   \\
				&       & \textcolor{blue}{4.2} (\textcolor{orange}{1$\times$})
				& \textcolor{blue}{6.9} (\textcolor{orange}{1.6$\times$})
				& \textcolor{blue}{12.6} (\textcolor{orange}{3.0$\times$})
				& \textcolor{blue}{39.7} (\textcolor{orange}{9.4$\times$})
				& \cellcolor{mytan}\textcolor{blue}{66.7} (\textcolor{orange}{15.9$\times$})\\
				\midrule
				\multirow{4}{*}{\centering MATH (4-shot)}
				& 256 & \textbf{33.5} & \underline{33.3}$^{*}$ & \underline{33.3} & 33.2 & \cellcolor{mytan}\underline{33.3} \\
				&       & \textcolor{blue}{8.7} (\textcolor{orange}{1$\times$})   
				& \textcolor{blue}{14.3} (\textcolor{orange}{1.6$\times$})
				& \textcolor{blue}{29.0} (\textcolor{orange}{3.3$\times$})
				& \textcolor{blue}{49.3} (\textcolor{orange}{5.7$\times$})
				& \cellcolor{mytan}\textcolor{blue}{71.0} (\textcolor{orange}{8.2$\times$})\\
				& 512 & \textbf{37.2} & \underline{36.3}$^{*}$ & 36.2 & 36.0 & \cellcolor{mytan}36.1 \\
				&       & \textcolor{blue}{6.2} (\textcolor{orange}{1$\times$})
				& \textcolor{blue}{10.1} (\textcolor{orange}{1.7$\times$})
				& \textcolor{blue}{18.6} (\textcolor{orange}{3.0$\times$})
				& \textcolor{blue}{43.2} (\textcolor{orange}{7.0$\times$})
				& \cellcolor{mytan}\textcolor{blue}{63.2} (\textcolor{orange}{10.2$\times$})\\
				\bottomrule
				\label{tab:LLaDA-tps}
			\end{tabular}
		\end{small}
	\end{center}
	\vspace{-0.4cm}
\end{table*}

\clearpage

\section{Additional Experiment Details.}
\subsection{More Experiments on LLaDA-Instruct}
\label{sec:More Experiments}

We report the accuracy and throughput results on LLaDA-Instruct in~\cref{tab:LLaDA-tps}. Our method achieves 6.4$\times$–20.6$\times$ speedup across all tasks. In comparison, the second-best method, Fast-dLLM, achieves 4.4$\times$–10.0$\times$ speedup with slightly lower accuracy. Specifically, our method improves accuracy by 13.8\%–27.0\% on MBPP. We attribute this improvement to the proposed attenuation guided suffix modeling, which effectively suppresses interference from semantically weak, high-entropy masked tokens in the distant suffix region during block-wise generation.

\subsection{Inference Latency Experiments for dLLMs}
\label{sec:Inference Latency}
In addition, we compare Streaming-dLLM and several other baselines in terms of per sample inference latency. This metric better reflects user-perceived responsiveness in practical scenarios, highlighting the suitability of our method for real-time deployment~\cite{xu2023lgvit}. As shown in~\cref{tab:dream_t}, Streaming-dLLM achieves a 4.8$\times$–21.0$\times$ speedup across all benchmarks compared with the vanilla backbone. Relative to the strongest acceleration baseline, it provides an additional 1.5$\times$–4.5$\times$ speedup on tasks with a generation length of 512, while maintaining comparable or slightly improved accuracy.

We observe even faster speed gains on LLaDA-1.5, as reported in~\cref{tab:llada_1.5_t}. In particular, Streaming-dLLM attains a 52.7$\times$ speedup on MBPP at generation length 512, markedly exceeding that of Fast-dLLM~\cite{wu2025fast}. Consistent improvements are also evident on LLaDA-Instruct of~\cref{tab:llada_t}, where our method achieves 5.3$\times$–64.1$\times$ speedup across all tasks. By comparison, the second-best method, Fast-dLLM, reaches only 3.9$\times$–9.9$\times$ speedup with slightly lower accuracy.

Overall, these results demonstrate that Streaming-dLLM consistently improves inference efficiency across different dLLM backbones and tasks without sacrificing generation quality. The substantial gains in both throughput and per sample latency highlight its practicality for real-world deployment, particularly in latency-sensitive and interactive applications.

\begin{table*}[ht]
	\centering
	\caption{
		Comparison of Dream-Base suite performance across four benchmarks at different generation lengths (256 and 512).Each cell reports accuracy (top) and inference latency with relative speedup over the Dream baseline (bottom; blue: s, orange: speedup). Results in bold indicate the highest score, while those underlined denote the second-best performance for each method. `\textsuperscript{*}' marks results we reproduced using the official implementation, as they were not reported in the original papers.}
	\renewcommand{\arraystretch}{1.16}
	\setlength{\tabcolsep}{6pt}
	\begin{center}
		\begin{small}
			\begin{tabular}{lcccccc}
				\toprule
				Benchmark & Gen Length & Dream & dKV-Cache & Prefix-Cache & Fast-dLLM & Ours  \\
				\midrule
				\multirow{4}{*}{\centering HumanEval (0-shot)}
				& 256 & 49.4 & 48.2$^{*}$ & \underline{53.7} & \textbf{54.3} & \cellcolor{mytan}\textbf{54.3} \\
				&       & \textcolor{green!60!black}{11.9} (\textcolor{orange}{1$\times$})
				& \textcolor{green!60!black}{11.9} (\textcolor{orange}{1.0$\times$})
				& \textcolor{green!60!black}{8.0} (\textcolor{orange}{1.5$\times$})
				& \textcolor{green!60!black}{4.6} (\textcolor{orange}{2.6$\times$})
				& \cellcolor{mytan}\textcolor{green!60!black}{2.5} (\textcolor{orange}{4.8$\times$})\\
				& 512 & 54.3 & 49.4$^{*}$ & \textbf{54.9} & 54.3 & \cellcolor{mytan}\underline{54.6} \\
				&       & \textcolor{green!60!black}{34.9} (\textcolor{orange}{1$\times$})
				& \textcolor{green!60!black}{32.5} (\textcolor{orange}{1.1$\times$})
				& \textcolor{green!60!black}{19.5} (\textcolor{orange}{1.8$\times$})
				& \textcolor{green!60!black}{12.1} (\textcolor{orange}{2.9$\times$})
				& \cellcolor{mytan}\textcolor{green!60!black}{3.1} (\textcolor{orange}{11.3$\times$})\\
				\midrule
				\multirow{4}{*}{\centering GSM8K-CoT (5-shot)}
				& 256 & \textbf{74.8} & 73.6$^{*}$ & \underline{74.0} & 73.5 & \cellcolor{mytan}\underline{74.0} \\
				&       & \textcolor{green!60!black}{18.6} (\textcolor{orange}{1$\times$})
				& \textcolor{green!60!black}{15.0} (\textcolor{orange}{1.2$\times$})
				& \textcolor{green!60!black}{8.1} (\textcolor{orange}{2.3$\times$})
				& \textcolor{green!60!black}{5.3} (\textcolor{orange}{3.5$\times$})
				& \cellcolor{mytan}\textcolor{green!60!black}{2.1} (\textcolor{orange}{8.9$\times$})\\
				& 512 & \underline{74.2} & 71.6$^{*}$ & \underline{74.2} & 74.1 & \cellcolor{mytan}\textbf{74.7} \\
				&       & \textcolor{green!60!black}{50.3} (\textcolor{orange}{1$\times$})
				& \textcolor{green!60!black}{39.9} (\textcolor{orange}{1.3$\times$})
				& \textcolor{green!60!black}{21.6} (\textcolor{orange}{2.3$\times$})
				& \textcolor{green!60!black}{10.7} (\textcolor{orange}{4.7$\times$})
				& \cellcolor{mytan}\textcolor{green!60!black}{2.4} (\textcolor{orange}{21.0$\times$})\\
				\midrule
				\multirow{4}{*}{\centering MBPP (3-shot)}
				& 256 & \textbf{56.6} & 54.0$^{*}$ & 53.2 & \underline{56.4} & \cellcolor{mytan}\underline{56.4} \\ 
				&       & \textcolor{green!60!black}{23.3} (\textcolor{orange}{1$\times$})
				& \textcolor{green!60!black}{17.4} (\textcolor{orange}{1.3$\times$})
				& \textcolor{green!60!black}{7.7} (\textcolor{orange}{3.0$\times$})
				& \textcolor{green!60!black}{3.8} (\textcolor{orange}{6.1$\times$})
				& \cellcolor{mytan}\textcolor{green!60!black}{3.3} (\textcolor{orange}{7.1$\times$})\\
				& 512 & \underline{55.6} & 53.0$^{*}$ & 53.8 & 55.2 & \cellcolor{mytan}\textbf{55.8}   \\
				&       & \textcolor{green!60!black}{58.6} (\textcolor{orange}{1$\times$})
				& \textcolor{green!60!black}{44.0} (\textcolor{orange}{1.3$\times$})
				& \textcolor{green!60!black}{20.9} (\textcolor{orange}{2.8$\times$})
				& \textcolor{green!60!black}{8.1} (\textcolor{orange}{7.2$\times$})
				& \cellcolor{mytan}\textcolor{green!60!black}{5.6} (\textcolor{orange}{10.5$\times$})\\
				\midrule
				\multirow{4}{*}{\centering MATH (4-shot)}
				& 256 & \textbf{38.4} & 36.8$^{*}$ & 36.8 & \underline{37.6} & \cellcolor{mytan}\underline{37.6} \\
				&       & \textcolor{green!60!black}{24.5} (\textcolor{orange}{1$\times$})
				& \textcolor{green!60!black}{17.5} (\textcolor{orange}{1.4$\times$})
				& \textcolor{green!60!black}{7.9} (\textcolor{orange}{3.1$\times$})
				& \textcolor{green!60!black}{4.2} (\textcolor{orange}{5.8$\times$})
				& \cellcolor{mytan}\textcolor{green!60!black}{3.2} (\textcolor{orange}{7.7$\times$})\\
				& 512 & \textbf{39.8} & 38.5$^{*}$ & 38.0 & 39.3 & \cellcolor{mytan}\underline{39.4} \\
				&       & \textcolor{green!60!black}{59.3} (\textcolor{orange}{1$\times$})
				& \textcolor{green!60!black}{44.1} (\textcolor{orange}{1.3$\times$})
				& \textcolor{green!60!black}{20.9} (\textcolor{orange}{2.8$\times$})
				& \textcolor{green!60!black}{9.4} (\textcolor{orange}{6.3$\times$})
				& \cellcolor{mytan}\textcolor{green!60!black}{5.2} (\textcolor{orange}{11.4$\times$})\\
				\bottomrule
				\label{tab:dream_t}
			\end{tabular}
		\end{small}
	\end{center}
	\vspace{-0.4cm}
\end{table*}

\clearpage
\begin{table*}[!ht]
	\centering
	\caption{
		Comparison of LLaDA-1.5 suite performance across four benchmarks at different generation lengths (256 and 512).Each cell reports accuracy (top) and inference latency with relative speedup over the LLaDA-1.5 baseline (bottom; blue: s, orange: speedup). Results in bold indicate the highest score, while those underlined denote the second-best performance for each method. `\textsuperscript{*}' marks results we reproduced using the official implementation, as they were not reported in the original papers.}
	\renewcommand{\arraystretch}{1.16}
	\setlength{\tabcolsep}{6pt}
	\begin{center}
		\begin{small}
			\begin{tabular}{lcccccc}
				\toprule
				Benchmark & Gen Length & LLaDA-1.5 & dKV-Cache & Prefix-Cache & Fast-dLLM & Ours  \\
				\midrule
				\multirow{4}{*}{\centering HumanEval (0-shot)}
				& 256 & \textbf{43.9}$^{*}$ & \underline{40.2}$^{*}$ & 38.4$^{*}$ & 37.2$^{*}$ & \cellcolor{mytan}39.0 \\
				&       & \textcolor{green!60!black}{12.7} (\textcolor{orange}{1$\times$})
				& \textcolor{green!60!black}{12.7} (\textcolor{orange}{1.0$\times$})
				& \textcolor{green!60!black}{7.72} (\textcolor{orange}{1.6$\times$})
				& \textcolor{green!60!black}{4.2} (\textcolor{orange}{3.0$\times$})
				& \cellcolor{mytan}\textcolor{green!60!black}{2.3} (\textcolor{orange}{5.5$\times$})\\
				& 512 & \textbf{40.5}$^{*}$ & \underline{40.2}$^{*}$ & 37.8$^{*}$ & 39.8$^{*}$ & \cellcolor{mytan}\underline{40.2} \\
				&       & \textcolor{green!60!black}{38.2} (\textcolor{orange}{1$\times$})
				& \textcolor{green!60!black}{32.4} (\textcolor{orange}{1.2$\times$})
				& \textcolor{green!60!black}{21.2} (\textcolor{orange}{1.8$\times$})
				& \textcolor{green!60!black}{7.6} (\textcolor{orange}{5.0$\times$})
				& \cellcolor{mytan}\textcolor{green!60!black}{2.9} (\textcolor{orange}{13.2$\times$})\\
				\midrule
				\multirow{4}{*}{\centering GSM8K (5-shot)}
				& 256 & 80.5$^{*}$ & \underline{80.7}$^{*}$ & 80.6$^{*}$ & \underline{80.7} & \cellcolor{mytan}\textbf{80.8} \\
				&       & \textcolor{green!60!black}{34.5} (\textcolor{orange}{1$\times$}) 
				& \textcolor{green!60!black}{20.1} (\textcolor{orange}{1.7$\times$})
				& \textcolor{green!60!black}{8.9} (\textcolor{orange}{3.9$\times$}) 
				& \textcolor{green!60!black}{4.4} (\textcolor{orange}{7.8$\times$}) 
				& \cellcolor{mytan}\textcolor{green!60!black}{2.0} (\textcolor{orange}{17.3$\times$}) \\
				& 512 & 81.0$^{*}$ & \textbf{81.3}$^{*}$ & 81.0$^{*}$ & 80.4 & \cellcolor{mytan}\underline{81.2} \\
				&       & \textcolor{green!60!black}{81.8} (\textcolor{orange}{1$\times$}) 
				& \textcolor{green!60!black}{48.1} (\textcolor{orange}{1.7$\times$}) 
				& \textcolor{green!60!black}{24.3} (\textcolor{orange}{3.4$\times$})
				& \textcolor{green!60!black}{7.8} (\textcolor{orange}{10.5$\times$}) 
				& \cellcolor{mytan}\textcolor{green!60!black}{2.2} (\textcolor{orange}{37.2$\times$}) \\
				\midrule
				\multirow{4}{*}{\centering MBPP (3-shot)}
				& 256 & \underline{38.0}$^{*}$ & \textbf{38.2}$^{*}$ & 37.8$^{*}$ & 37.6$^{*}$ & \cellcolor{mytan}37.8 \\ 
				&       & \textcolor{green!60!black}{28.7} (\textcolor{orange}{1$\times$})   
				& \textcolor{green!60!black}{17.7} (\textcolor{orange}{1.6$\times$})
				& \textcolor{green!60!black}{8.5} (\textcolor{orange}{3.4$\times$})
				& \textcolor{green!60!black}{2.3} (\textcolor{orange}{12.5$\times$})
				& \cellcolor{mytan}\textcolor{green!60!black}{1.2} (\textcolor{orange}{23.9$\times$})\\
				& 512 & \underline{38.2}$^{*}$ & 38.1$^{*}$ & 38.0$^{*}$ & 38.1$^{*}$ & \cellcolor{mytan}\textbf{38.4}   \\
				&       & \textcolor{green!60!black}{68.5} (\textcolor{orange}{1$\times$})
				& \textcolor{green!60!black}{43.0} (\textcolor{orange}{1.6$\times$})
				& \textcolor{green!60!black}{23.1} (\textcolor{orange}{3.0$\times$})
				& \textcolor{green!60!black}{4.1} (\textcolor{orange}{16.7$\times$})
				& \cellcolor{mytan}\textcolor{green!60!black}{1.3} (\textcolor{orange}{52.7$\times$})\\
				\midrule
				\multirow{4}{*}{\centering MATH (4-shot)}
				& 256 & \underline{32.7}$^{*}$ & 31.8$^{*}$ & 32.5$^{*}$ & 32.6 & \cellcolor{mytan}\textbf{33.7} \\
				&       & \textcolor{green!60!black}{27.9} (\textcolor{orange}{1$\times$})   
				& \textcolor{green!60!black}{17.6} (\textcolor{orange}{1.6$\times$})
				& \textcolor{green!60!black}{8.5} (\textcolor{orange}{3.3$\times$})
				& \textcolor{green!60!black}{4.7} (\textcolor{orange}{5.9$\times$})
				& \cellcolor{mytan}\textcolor{green!60!black}{2.7} (\textcolor{orange}{10.3$\times$})\\
				& 512 & \textbf{37.1}$^{*}$ & \underline{35.1}$^{*}$ & 35.0$^{*}$ & \underline{35.1} & \cellcolor{mytan}\underline{35.1} \\
				&       & \textcolor{green!60!black}{66.3} (\textcolor{orange}{1$\times$})
				& \textcolor{green!60!black}{42.1} (\textcolor{orange}{1.6$\times$})
				& \textcolor{green!60!black}{22.9} (\textcolor{orange}{2.9$\times$})
				& \textcolor{green!60!black}{8.4} (\textcolor{orange}{7.9$\times$})
				& \cellcolor{mytan}\textcolor{green!60!black}{4.6} (\textcolor{orange}{14.4$\times$})\\
				\bottomrule
				\label{tab:llada_1.5_t}
			\end{tabular}
		\end{small}
	\end{center}
	\vspace{-0.4cm}
\end{table*}
\begin{table*}[!ht]
	\centering
	\caption{
		Comparison of LLaDA-Instruct suite performance across four benchmarks at different generation lengths (256 and 512).Each cell reports accuracy (top) and inference latency with relative speedup over the LLaDA baseline (bottom; blue: s, orange: speedup). Results in bold indicate the highest score, while those underlined denote the second-best performance for each method. `\textsuperscript{*}' marks results we reproduced using the official implementation, as they were not reported in the original papers.}
	\renewcommand{\arraystretch}{1.16}
	\setlength{\tabcolsep}{6pt}
	\begin{center}
		\begin{small}
			\begin{tabular}{lcccccc}
				\toprule
				Benchmark & Gen Length & LLaDA & dKV-Cache & Prefix-Cache & Fast-dLLM & Ours  \\
				\midrule
				\multirow{4}{*}{\centering HumanEval (0-shot)}
				& 256 & 41.5 & 39.6$^{*}$ & 42.7 & \underline{43.3} & \cellcolor{mytan}\textbf{43.4} \\
				&       & \textcolor{green!60!black}{13.3} (\textcolor{orange}{1$\times$})
				& \textcolor{green!60!black}{12.9} (\textcolor{orange}{1.0$\times$})
				& \textcolor{green!60!black}{7.9} (\textcolor{orange}{1.7$\times$})
				& \textcolor{green!60!black}{3.4} (\textcolor{orange}{3.9$\times$})
				& \cellcolor{mytan}\textcolor{green!60!black}{2.5} (\textcolor{orange}{5.3$\times$})\\
				& 512 & 43.9 & \underline{45.1}$^{*}$ & \textbf{45.7} & 44.5 & \cellcolor{mytan}44.5 \\
				&       & \textcolor{green!60!black}{39.4} (\textcolor{orange}{1$\times$})
				& \textcolor{green!60!black}{32.9} (\textcolor{orange}{1.2$\times$})
				& \textcolor{green!60!black}{21.4} (\textcolor{orange}{1.8$\times$})
				& \textcolor{green!60!black}{9.0} (\textcolor{orange}{4.4$\times$})
				& \cellcolor{mytan}\textcolor{green!60!black}{5.2} (\textcolor{orange}{7.6$\times$})\\
				\midrule
				\multirow{4}{*}{\centering GSM8K (5-shot)}
				& 256 & \underline{79.3} & 76.5$^{*}$ & \textbf{79.5} & 78.5 & \cellcolor{mytan}78.5 \\
				&       & \textcolor{green!60!black}{35.5} (\textcolor{orange}{1$\times$})
				& \textcolor{green!60!black}{20.2} (\textcolor{orange}{1.8$\times$})
				& \textcolor{green!60!black}{9.0} (\textcolor{orange}{3.9$\times$})
				& \textcolor{green!60!black}{4.6} (\textcolor{orange}{7.7$\times$})
				& \cellcolor{mytan}\textcolor{green!60!black}{2.2} (\textcolor{orange}{16.1$\times$})\\
				& 512 & 77.5 & \underline{77.6}$^{*}$ & 77.0 & 77.2 & \cellcolor{mytan}\textbf{78.7} \\
				&       & \textcolor{green!60!black}{84.1} (\textcolor{orange}{1$\times$})
				& \textcolor{green!60!black}{49.0} (\textcolor{orange}{1.7$\times$})
				& \textcolor{green!60!black}{24.7} (\textcolor{orange}{3.4$\times$})
				& \textcolor{green!60!black}{8.5} (\textcolor{orange}{9.9$\times$})
				& \cellcolor{mytan}\textcolor{green!60!black}{2.3} (\textcolor{orange}{36.6$\times$})\\
				\midrule
				\multirow{4}{*}{\centering MBPP (3-shot)}
				& 256 & 29.4 & 26.2$^{*}$ & \underline{29.6} & 28.2 & \cellcolor{mytan}\textbf{42.0} \\ 
				&       & \textcolor{green!60!black}{29.4} (\textcolor{orange}{1$\times$})
				& \textcolor{green!60!black}{17.6} (\textcolor{orange}{1.7$\times$})
				& \textcolor{green!60!black}{8.6} (\textcolor{orange}{3.4$\times$})
				& \textcolor{green!60!black}{3.8} (\textcolor{orange}{7.7$\times$})
				& \cellcolor{mytan}\textcolor{green!60!black}{0.9} (\textcolor{orange}{32.7$\times$})\\
				& 512 & \underline{14.8} & 14.6$^{*}$ & 13.4 & 13.8 & \cellcolor{mytan}\textbf{40.8}   \\
				&       & \textcolor{green!60!black}{70.5} (\textcolor{orange}{1$\times$})
				& \textcolor{green!60!black}{43.3} (\textcolor{orange}{1.6$\times$})
				& \textcolor{green!60!black}{23.6} (\textcolor{orange}{3.0$\times$})
				& \textcolor{green!60!black}{7.6} (\textcolor{orange}{9.3$\times$})
				& \cellcolor{mytan}\textcolor{green!60!black}{1.1} (\textcolor{orange}{64.1$\times$})\\
				\midrule
				\multirow{4}{*}{\centering MATH (4-shot)}
				& 256 & \textbf{33.5} & \underline{33.3}$^{*}$ & \underline{33.3} & 33.2 & \cellcolor{mytan}\underline{33.3} \\
				&       & \textcolor{green!60!black}{28.6} (\textcolor{orange}{1$\times$})
				& \textcolor{green!60!black}{17.4} (\textcolor{orange}{1.6$\times$})
				& \textcolor{green!60!black}{8.6} (\textcolor{orange}{3.3$\times$})
				& \textcolor{green!60!black}{5.1} (\textcolor{orange}{5.6$\times$})
				& \cellcolor{mytan}\textcolor{green!60!black}{3.5} (\textcolor{orange}{8.2$\times$})\\
				& 512 & \textbf{37.2} & \underline{36.3}$^{*}$ & 36.2 & 36.0 & \cellcolor{mytan}36.1 \\
				&       & \textcolor{green!60!black}{68.8} (\textcolor{orange}{1$\times$})
				& \textcolor{green!60!black}{42.7} (\textcolor{orange}{1.6$\times$})
				& \textcolor{green!60!black}{23.3} (\textcolor{orange}{3.0$\times$})
				& \textcolor{green!60!black}{10.0} (\textcolor{orange}{6.9$\times$})
				& \cellcolor{mytan}\textcolor{green!60!black}{6.8} (\textcolor{orange}{10.1$\times$})\\
				\bottomrule
				\label{tab:llada_t}
			\end{tabular}
		\end{small}
	\end{center}
	\vspace{-0.4cm}
\end{table*}

\clearpage

\subsection{Optimization and Schedule.}
\label{sec:Optimization and Schedule}
We provide detailed configurations for evaluation on Humaneval, GSM8K, MBPP, and MATH, as summarized in Table~\ref{tab:hyperparameters}. For each benchmark and backbone (Dream, LLaDA, and LLaDA-1.5), we report the generation length, sliding window size, initial confidence threshold $\tau_0$, decay factor $\alpha$, and block size used in our experiments. For all models, the block size is set to 32 across datasets, while other hyperparameters vary by benchmark and generation length. These settings are consistently applied for runs reported in the main paper.

\begin{table}[ht]
	\centering
	\caption{Configurations for different dataset.}
	\label{tab:hyperparameters}
	\begin{tabular}{llccccc}
		\toprule
		\textbf{Model} & \textbf{Benchmark} & \textbf{Gen Length} &  \textbf{sliding window size} & \textbf{$\tau_0$} & \textbf{$\alpha$} &  \textbf{block\_size} \\
		\midrule
		\multirow{9}{*}{\textbf{Dream}} & \multirow{2}{*}{Humaneval (0-shot)} & 256 & 192 & 0.9 & 0.7 & 32 \\
		& & 512 & 128 & 0.9 & 0.4 & 32\\
		\cmidrule(l){2-7}
		& \multirow{2}{*}{GSM8K-CoT (5-shot)} & 256 & 32 & 0.9 & 0.3 & 32\\
		& & 512 & 32 & 0.9 & 0.3 & 32\\
		\cmidrule(l){2-7}
		& \multirow{2}{*}{MBPP (3-shot)} & 256 & 192 & 0.9 & 0.3 & 32\\
		& & 512 & 192 & 0.9 & 0.6 & 32\\
		\cmidrule(l){2-7}
		& \multirow{2}{*}{MATH (4-shot)} & 256 & 32 & 0.9 & 0.1 & 32\\
		& & 512 & 32 & 0.9 & 0.3 & 32\\
		\midrule
		\multirow{9}{*}{\textbf{LLaDA}} & \multirow{2}{*}{Humaneval (0-shot)} & 256 & 192 & 0.9 & 0.3 & 32\\
		& & 512 & 256 & 0.9 & 0.4 & 32\\
		\cmidrule(l){2-7}
		& \multirow{2}{*}{GSM8K (5-shot))} & 256 & 96 & 0.9 & 0.3 & 32\\
		& & 512 & 96 & 0.9 & 0.3 & 32\\
		\cmidrule(l){2-7}
		& \multirow{2}{*}{MBPP (3-shot)} & 256 & 32 & 0.9 & 0.3 & 32\\
		& & 512 & 32 & 0.9 & 0.3 & 32\\
		\cmidrule(l){2-7}
		& \multirow{2}{*}{MATH (4-shot)} & 256 & 128 & 0.9 & 0.3 & 32\\
		& & 512 & 256 & 0.9 & 0.2 & 32\\
		\midrule
		\multirow{9}{*}{\textbf{LLaDA-1.5}} & \multirow{2}{*}{Humaneval (0-shot)} & 256 & 96 & 0.9 & 0.3 & 32\\
		& & 512 & 96 & 0.9 & 0.4 & 32 \\
		\cmidrule(l){2-7}
		& \multirow{2}{*}{GSM8K (5-shot)} & 256 & 96 & 0.9 & 0.4 & 32\\
		& & 512 & 128 & 0.9 & 0.6 & 32\\
		\cmidrule(l){2-7}
		& \multirow{2}{*}{MBPP (3-shot)} & 256 & 96 & 0.9 & 0.3 & 32\\
		& & 512 & 96 & 0.9 & 0.3 & 32\\
		\cmidrule(l){2-7}
		& \multirow{2}{*}{MATH (4-shot)} & 256 & 96 & 0.9 & 0.4 & 32\\
		& & 512 & 192 & 0.9 & 0.3 & 32\\
		\bottomrule
	\end{tabular}
	\vspace{-0.4cm}
\end{table}

\subsection{Additional Analysis for Generation Length.}

\begin{table}[ht]
	\caption{Impact of generation length on Accuracy and Speedup for LLaDA on GSM8K (5-Shot).}
	\label{tab:ablation5}
	\begin{center}
		\begin{small}
			\begin{tabular}{lccc}
				\toprule
				\textbf{Model} & \textbf{512} & \textbf{1024} & \textbf{2048}  \\
				\midrule
				\multirow{2}{*}{\centering LLaDA}  & 77.5 & 77.6 & 76.8    \\ 
				& \textcolor{blue}{3.2} (\textcolor{orange}{1$\times$})
				& \textcolor{blue}{1.1} (\textcolor{orange}{1$\times$}) 
				& \textcolor{blue}{0.4} (\textcolor{orange}{1$\times$}) \\
				\multirow{2}{*}{\centering Fast-dLLM}  & 77.2 & 75.3 & 76.4    \\
				& \textcolor{blue}{32.1} (\textcolor{orange}{10.0$\times$})
				& \textcolor{blue}{15.0} (\textcolor{orange}{13.6$\times$}) 
				& \textcolor{blue}{6.5} (\textcolor{orange}{16.3$\times$}) \\
				\multirow{2}{*}{\centering Ours}  & 78.7 & 78.3 & 78.7     \\
				& \textcolor{blue}{65.9} (\textcolor{orange}{20.6$\times$})
				& \textcolor{blue}{64.7} (\textcolor{orange}{58.8$\times$}) 
				& \textcolor{blue}{67.1} (\textcolor{orange}{167.8$\times$}) \\
				\bottomrule
			\end{tabular}
		\end{small}
	\end{center}
	\vspace{-0.4cm}
\end{table}

Table~\ref{tab:ablation5} reports the accuracy and speedup of LLaDA on GSM8K (5-shot) under different generation lengths (512, 1024, and 2048). For each model, the top row shows the accuracy, while the bottom row reports the throughput (tokens per second) in blue and the corresponding speedup relative to the vanilla backbone in orange. The results indicate that our method consistently maintains high accuracy across all generation lengths while achieving substantially larger speedup compared with both LLaDA and Fast-dLLM, especially as the generation length increases.

\end{document}